\documentclass[10pt,twocolumn,letterpaper]{article}

\usepackage{iccv}
\usepackage{times}
\usepackage{epsfig}
\usepackage{graphicx}
\usepackage{amsmath}
\usepackage{amssymb}
\usepackage{blindtext}
\usepackage{graphicx}
\usepackage{caption}
\usepackage{subcaption}
\usepackage{enumitem}
\usepackage{booktabs}
\usepackage{placeins}
\usepackage{multirow}
\usepackage[pagebackref=true,breaklinks=true,letterpaper=true,colorlinks,bookmarks=false]{hyperref}
\usepackage{cleveref}
\usepackage{array}
\newcolumntype{P}[1]{>{\centering\arraybackslash}p{#1}}
\usepackage{dblfloatfix}
\usepackage[dvipsnames]{xcolor}
\usepackage{hyperref}

\usepackage[pagebackref=true,breaklinks=true,letterpaper=true,colorlinks,bookmarks=false]{hyperref}
\usepackage[accsupp]{axessibility} 
\usepackage[symbol]{footmisc}

\iccvfinalcopy 

\def\iccvPaperID{4449} 

\begin{document}

\title{Perceptual Artifacts Localization for Image Synthesis Tasks}

\author{
\begin{tabular}{ccccccc}
Lingzhi Zhang\textsuperscript{$\scalebox{0.7}{$\bigstar$}$$\scalebox{0.7}{$\spadesuit$}$ 1,2} & 
Zhengjie Xu\textsuperscript{$\scalebox{0.7}{$\bigstar$}$ 2} & 
Connelly Barnes\textsuperscript{1} & 
Yuqian Zhou\textsuperscript{1} & 
Qing Liu\textsuperscript{1} \\
He Zhang\textsuperscript{1} & 
Sohrab Amirghodsi\textsuperscript{1} & 
Zhe Lin\textsuperscript{1} &
Eli Shechtman\textsuperscript{1} & 
Jianbo Shi\textsuperscript{2} \\
\end{tabular}\\[3ex]
\textsuperscript{1}Adobe Inc. \quad
\textsuperscript{2}University of Pennsylvania\\
}

\twocolumn[{%
\maketitle
\begin{center}
    \centering
    \includegraphics[trim=0in 0.6in 1.6in 0in, clip,width=\textwidth]{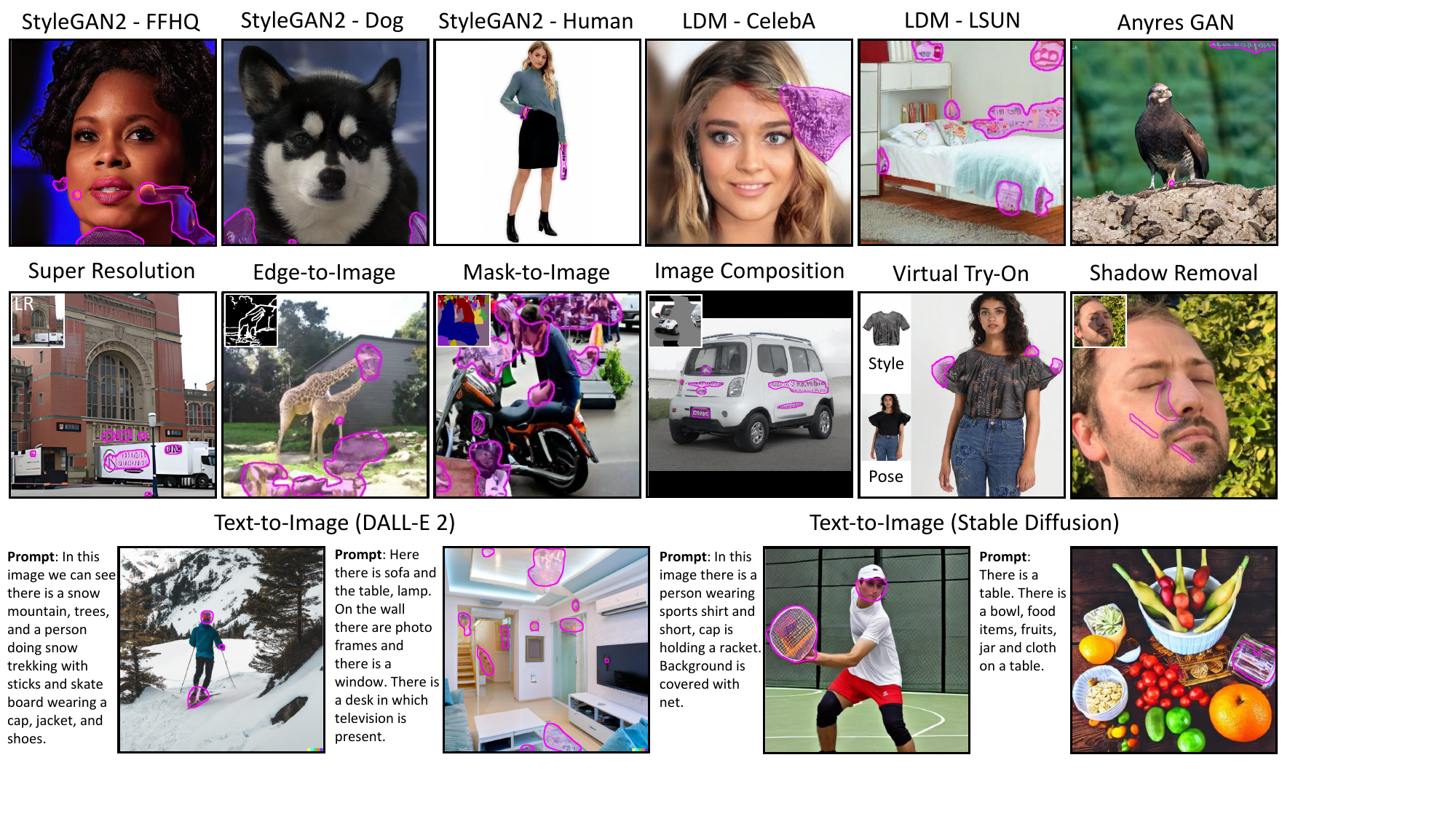}
    \captionof{figure}[Teaser]{The visualization of predicted perceptual artifacts localization on ten image synthesis tasks. The first row contains unconditionally generated images from StyleGAN2 \cite{karras2020analyzing}, Latent Diffusion Model (LDM) \cite{rombach2022high}, and Anyres GAN \cite{chai2022any}. The second row shows the results on types of conditional generated images, including super-resolution with Real-ESRGAN \cite{wang2021real}, edge-to-image with PITI \cite{wang2022pretraining}, mask-to-image with PITI \cite{wang2022pretraining}, image latent composition \cite{chai2021using}, virtual try-on \cite{fele2022c}, and portrait shadow removal \cite{zhang2020portrait}. The conditional inputs are placed at the top left of the images. In the last row, we show predictions on the text-to-image outputs from DALL-E 2 \cite{ramesh2022hierarchical} and Stable Diffusion \cite{rombach2022high}. }
    \label{fig:teaser}
\end{center}%
}]

\footnotetext{$\bigstar$ indicates equal contribution. $\spadesuit$ work done when Lingzhi is a graduate student at University of Pennsylvania.}

\begin{abstract}

Recent advancements in deep generative models have facilitated the creation of photo-realistic images across various tasks. However, these generated images often exhibit perceptual artifacts in specific regions, necessitating manual correction. In this study, we present a comprehensive empirical examination of Perceptual Artifacts Localization (PAL) spanning diverse image synthesis endeavors. We introduce a novel dataset comprising 10,168 generated images, each annotated with per-pixel perceptual artifact labels across ten synthesis tasks. A segmentation model, trained on our proposed dataset, effectively localizes artifacts across a range of tasks. Additionally, we illustrate its proficiency in adapting to previously unseen models using minimal training samples. We further propose an innovative zoom-in inpainting pipeline that seamlessly rectifies perceptual artifacts in the generated images. Through our experimental analyses, we elucidate several practical downstream applications, such as automated artifact rectification, non-referential image quality evaluation, and abnormal region detection in images. The dataset and code are released here: {\href{https://owenzlz.github.io/PAL4VST/}{https://owenzlz.github.io/PAL4VST}}

\end{abstract}

\section{Introduction}
\label{sec:intro}

Generative models have made significant progress in a myriad of image synthesis tasks, including unconditional generation \cite{brock2018large, kingma2018glow, karras2020analyzing, karras2021alias, dhariwal2021diffusion}, image inpainting \cite{zhao2021large, suvorov2022resolution, lugmayr2022repaint, li2022mat, zheng2022image, zhang2022inpainting}, image-to-image translation \cite{park2019semantic, richardson2021encoding, su2022dual, saharia2022palette, wang2022pretraining}, and text-to-image synthesis \cite{ding2022cogview2, yu2022scaling, nichol2021glide, ramesh2022hierarchical, rombach2022high, saharia2022photorealistic, balaji2022ediffi}, among others. However, even cutting-edge models occasionally generate implausible content or display unpleasant artifacts in specific regions of the image, which we refer to as perceptual artifacts. These artifacts are easily detectable by the human eye. Therefore, in typical image editing processes, users often retouch generated images, masking and re-editing these regions to achieve perfection.

The manual retouching of perceptual artifacts is time-consuming and iterative. Such artifacts also pose challenges for generative models in achieving full automation in image synthesis, editing, or batch processing without human oversight. These challenges drive our exploration into the feasibility of training AI oracle models to identify and segment these perceptual artifacts. A successful implementation would present users with an automatically delineated mask of potential artifact areas, eliminating manual masking. Moreover, we could offer users the option to deploy established editing techniques, like inpainting, to these detected regions, thereby enhancing the automation of the retouching process.

Technically, the ideal goal is to generate a flawless image in a single pass. However, today's leading large-scale diffusion models often struggle to capture intricate details like subtle facial features, hands, and other object-specific nuances. While integrating more training data or using weighted loss might appear as potential solutions to these issues, they could compromise image quality in broader contexts. Until we achieve perfect single-pass outputs, automating the localization and refinement of perceptual artifacts stands as a promising direction to improve image synthesis quality.

To meet this objective, we've amassed a dataset of generated images, complemented with per-pixel artifact segmentation labels across a range of synthesis tasks. Using this dataset, we trained a segmentation model adept at localizing perceptual artifacts across various tasks. Our pretrained artifact detector showcases its versatility across multiple new models, adapting with enhanced accuracy even with limited training samples.

In conjunction with our artifact detection, we also unveil several practical applications. The foremost of these is the automatic refinement of artifacts in generated images using inpainting. However, it's observed that leading diffusion inpainting models, like DALL-E \cite{ramesh2022hierarchical} and Stable Diffusion \cite{rombach2022high}, sometimes falter in generating high-fidelity object details, such as facial features. We hypothesize this may stem from an unsuitable inpainting context. Consequently, we introduce a zoom-in inpainting pipeline, presenting a more apt input context before inpainting. This simple approach effectively mitigates challenges tied to object detail generation, without necessitating model training or alterations.

Our primary contributions include: 
\begin{itemize}[noitemsep]
  \item A novel high-quality dataset comprising 10,168 images with per-pixel artifact annotations from humans, spanning ten diverse image synthesis tasks.
  \item A segmentation model adept at localizing perceptual artifacts across multiple synthesis tasks. Our pretrained model exhibits a rapid adaptation capability to new techniques with minimal training examples.
  \item An novel zoom-in inpainting pipeline for the automated refinement of intricate details in generated images.
  \item Demonstrated applications of our artifact detector, which include: 1). automatic artifact refinement; 2). reference-free image quality evaluation; and 3). anomaly detection in natural images.
\end{itemize}

We will release the dataset and the code.

\section{Related Work}

\begin{figure*}[!t]
    \centering
    \includegraphics[trim=0.0in 4.9in 0.0in 0in, clip,width=\textwidth]{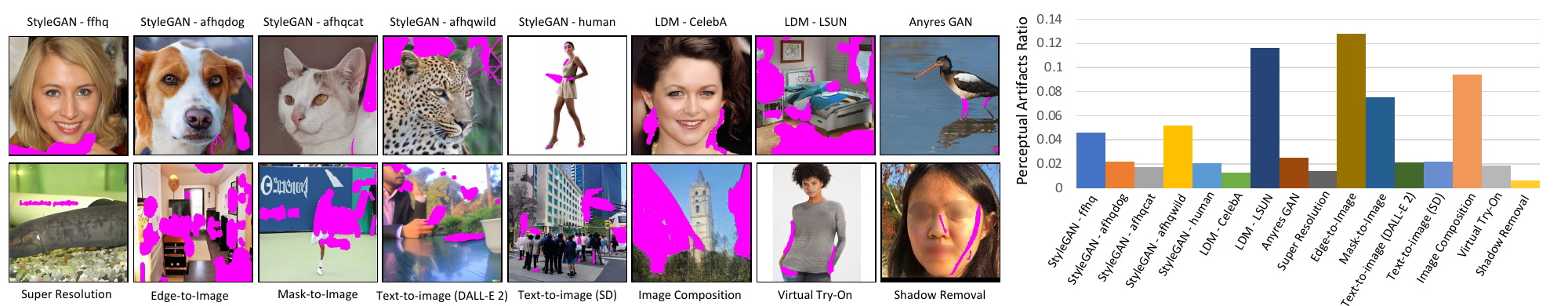}
    \caption{The left figure shows the raw labels, where the image order (left to right, and top to bottom) follows the order in the histogram. The histogram demonstrates the Perceptual Artifacts Ratio computed from human labels for different tasks and domains. }
    \label{fig:dataset}
\end{figure*}

\begin{figure}[!h]
    \centering
    \includegraphics[trim=0.0in 5.3in 7.15in 0in, clip,width=0.47\textwidth]{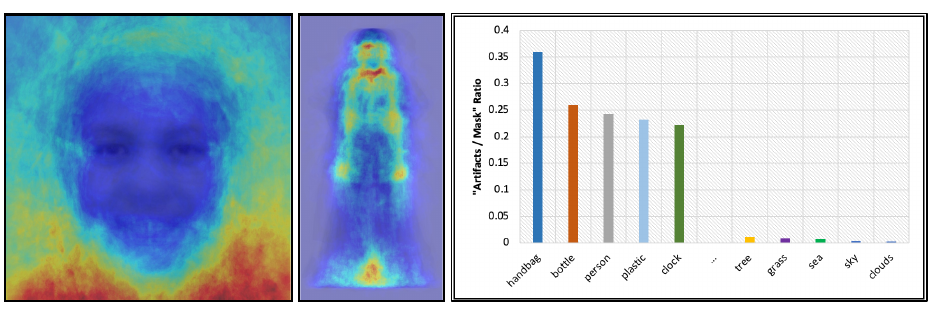}
    \caption{Visualization of the distribution of perceptual artifacts: On the left, for the StyleGAN2-generated \cite{karras2020analyzing} images, we observe that perceptual artifacts predominantly concentrate below the chin and around the neck region in facial images, and around the keypoint regions in full-body human images. On the right, we display the five COCO-stuff \cite{caesar2018coco} semantic classes that exhibit the highest and lowest amounts of artifacts for in-the-wild generated images. }
    \label{fig:artifacts_distribution}
\end{figure}

\textbf{Detecting Generated Images.} As generated images become increasingly photo-realistic, numerous studies \cite{Zhao_2021_CVPR, Liu_2021_CVPR, afchar2018mesonet, bayar2016deep, cozzolino2017recasting, mo2018fake, rossler2019faceforensics++, chai2020makes, nataraj2019detecting, barni2020cnn} have sought to automatically detect machine-generated images for forensic purposes. Given that new state-of-the-art generative models emerge frequently, an essential question arises: Can we train a model that generalizes to entirely unseen generative models? This query is explored in several works \cite{cozzolino2018forensictransfer, wang2020cnn, zhang2019detecting, xuan2019generalization}. Notably, several of these studies \cite{frank2020leveraging, wang2020cnn, liu2022detecting} have discerned that high-frequency details in both generated and real images can serve as valuable indicators, enabling the development of classifiers that can distinguish between fake and real images up to a certain extent. For example, Wang et al. \cite{wang2020cnn} demonstrated that classifiers, when trained on images generated by a specific GAN, can detect the majority of fake images stemming from other unencountered generative models. Chai et al. \cite{chai2020makes} devised a patch-based classifier with limited receptive fields, shedding light on the regions of fake images that are more transparently discerned. More recently, Ojha and Li et al. \cite{ojha2023towards} identified that an earlier classifier \cite{wang2019detecting} had an inclination to overfit to high-frequency noise present in GANs, leading it to erroneously categorize all diffusion-generated images as real. To rectify this, they incorporated frozen pretrained CLIP-ViT features \cite{radford2021learning}, deploying techniques like nearest neighbors or linear probing to strengthen the model's generalization capabilities. Bringing it closer to the context of our work, techniques like Grad-CAM \cite{selvaraju2017grad} can be employed to visualize the 'active' regions in a real-vs-fake classifier, explicitly highlighting the areas indicative of 'fakeness' from the \emph{perspective of artificial neural networks}.

\textbf{Localizing Editing Regions. } Beyond simply classifying images as real or generated, numerous research efforts have sought to localize the edited regions within the generated or edited images. For example, Wang et al. \cite{wang2019detecting} exploited Photoshop's scripting capabilities to automatically generate edited images. They then effectively trained a model to predict manipulated facial areas in photos. In the field of image inpainting, several studies \cite{wu2021iid, li2021noise, wang2021image} have demonstrated that high-frequency noise can be sufficiently informative to precisely segment inpainted areas, especially when training a model alongside the ground truth inpainting mask. While these studies bear some resemblance to our work, their main objective is to pinpoint systematic inconsistencies in generated images to distinguish fakes. In contrast, our research is centered on detecting and segmenting artifact areas that are noticeable to \emph{human perception}. We argue that localizing perceptual artifacts at a fine-grained level, rather than the entire edited region, can pave the way for enhanced generated quality. Notably, Zhang et al. \cite{zhang2022perceptual} focused on perceptual artifacts in inpainting tasks. Our methodology expands upon this, aiming at various synthesis tasks and their potential downstream applications.

\textbf{Improving Synthesis Quality During Inference.} Several previous works have introduced techniques to enhance image synthesis quality during inference time \cite{liu2020collaborative, brock2018large, dhariwal2021diffusion}, and our research aligns with this domain. For instance, the truncation trick, initially introduced in BigGAN \cite{brock2018large}, limits latent code sampling to a constrained space using a specific threshold. This method has been observed to result in improved visual quality of individual samples. Classifier guidance, as detailed in \cite{dhariwal2021diffusion} for diffusion-based models, suggests utilizing the gradient of a pretrained classifier to steer the diffusion process in image generation. This ensures that the generated images are accurately recognized by the classifier. Contrasting with these methods, our approach specifically targets the detection and refinement of perceptual artifact regions in images while striving to retain as much of the original content as possible. This differs from the 'hard' sample rejection seen in the truncation trick or the overarching image synthesis guidance provided by classifier guidance. Furthermore, our method neither relies on hyperparameters, such as the truncation threshold, nor requires gradient computation.


\section{Methods}
\label{sec:methods}

\subsection{Data Collection and Statistics}
\label{subsec:data_collection}

We collect our dataset by running inference using the pretrained models from ten different synthesis tasks. Within each task, we might run more than one model or checkpoint if the model, i,e. StyleGAN2 \cite{karras2020analyzing}, are trained on multiple domains. Overall, we collect 10,168 images with the per-pixel artifacts segmentation labels by human experts. Each image takes roughly one minute to label by a human expert, and thus, the entire dataset cost $\sim$170 hours of labor. We split the dataset into a train/test/val set divided as 80\%/10\%/10\%, respectively. 


\textbf{Statistics on Tasks.} The area of perceptual artifacts region highly depends on the complexity of image content, the nature of the task, and the performance of the synthesis model. As shown in the left of Figure \ref{fig:dataset}, we can see that some generated images receive larger marked artifacts regions than the others. We compute the Perceptual Artifacts Ratio (PAR), which is simply the labeled artifacts region divided by the image area, to quantify the levels of artifacts for each task. As shown in the histogram of Figure \ref{fig:dataset}, we can see that the tasks like Edge-to-Image \cite{wang2022pretraining}, Mask-to-Image \cite{wang2022pretraining}, LDM-LSUN \cite{rombach2022high}, and Image Composition \cite{chai2021using} have obviously larger PAR than the others, since the models are generating relatively complex in-the-wild visual content. 

\textbf{Statistics on Content.} Perceptual artifacts are more prevalent in certain object categories or semantic parts of images for two primary reasons. Firstly, visual content with significant variations is inherently challenging to generate. Secondly, human perceptual judgments tend to be more sensitive to specific regions. With these considerations, we embarked on a quantitative exploration of the distribution of perceptual artifacts in generated images. As depicted on the left of Figure \ref{fig:artifacts_distribution}, we calculated the average PAR heatmap for both face and human images. The heatmap for faces reveals that artifact-prone areas primarily fall around and beneath the chin. This is a region where StyleGAN2 \cite{karras2020analyzing} often endeavors to generate content with high variance—like microphones, necklaces, and clothing—but struggles to maintain high fidelity. The heatmap for human images points out that artifacts predominantly arise around specific human keypoints, such as the head, neck, hands, and feet—areas to which human perception is especially attuned. Regarding in-the-wild generated images, as seen on the right side of Figure \ref{fig:artifacts_distribution}, we discern that "object" regions typically exhibit a higher PAR compared to "stuff" regions.


\begin{table*}[!b]
    \begin{center}
     \resizebox{\textwidth}{!}{
      \begin{tabular}{l|c|c|c|c|c|c|c|c|c|c|c}
        \toprule 
        \textbf{Methods} & \ StyleGAN2 & \ LDMs & \ AnyRes & \ SR & \ Inpaint & \ E2I & \ M2I & \ T2I & \ Comp. & \ VTON & \ PSR  \\
        \midrule 
        CNNgenerates \cite{wang2020cnn} + Grad-CAM \cite{selvaraju2017grad} & 4.38 & 2.43 & 1.39 & 0.86 & 3.54 & 0.95 & 0.48 & 0.51 & 7.13 & 2.56 & 0.0 \\
        Patch Forensics \cite{chai2020makes} & 3.81 & 9.08 & 8.76 & 1.34 & 5.35 & 14.19 & 10.54 & 2.71 & 9.63 & 2.14 & 0.66 \\
        PAL4Inpaint \cite{zhang2022perceptual} & 6.12 & 0.98 & 1.03 & 0.81 & 42.07 & 0.86 & 0.31 & 0.51 & 14.42 & 15.94 & 0.0 \\
        \midrule
        Specialist Model (Ours) & 37.85 & \textbf{35.39} & 9.15 & \textbf{14.41} & \textbf{42.07} & 45.56 & 35.01 & \textbf{21.79} & 25.31 & 37.44 & \textbf{21.33} \\
        Unified Model (Ours) & \textbf{38.53} & 30.86 & \textbf{34.74} & 11.92 & 41.81 & \textbf{46.01} & \textbf{39.37} & 19.65 & \textbf{29.53} & \textbf{38.07} & 5.10 \\
        \bottomrule
      \end{tabular}
      }
      \caption{Quantitative mIoU ($\uparrow$) evaluation of perceptual artifacts segmentation on 10 image synthesis tasks. We use the following brevity to indicate the tasks: LDMs $\rightarrow$ Latent Diffusion Models \cite{rombach2022high}, SR $\rightarrow$ Super Resolution \cite{wang2021real}, E2I $\rightarrow$ Edge-to-Image \cite{wang2022pretraining}, M2I $\rightarrow$ Mask-to-Image \cite{wang2022pretraining}, T2I $\rightarrow$ Text-to-Image \cite{rombach2022high, ramesh2022hierarchical}, Comp. $\rightarrow$ Image Composition \cite{chai2021using}, VTON $\rightarrow$ Virtual Try-On \cite{fele2022c}, PSR $\rightarrow$ Portrait Shadow Removal \cite{zhang2020portrait}, and finally U.M. $\rightarrow$ Unified Model. }
      \label{tab:unified_vs_specialist}
    \end{center}
\end{table*}


\subsection{Segmenting Perceptual Artifacts} 

\textbf{Training Segmentation Models.} We formulate the localization of perceptual artifacts as a binary semantic segmentation problem. To train a single unified model for detecting generic perceptual artifacts, we use data collected from all tasks. During training, we adopt random cropping and horizontal flipping to augment the dataset. Our model is implemented with a Swin-T \cite{liu2021swin} backbone, where UperNet \cite{wang2020deep} serves as the main head and FCN \cite{long2015fully} as the auxiliary head. We train the model using a cross entropy loss and optimize it with the AdamW \cite{loshchilov2017decoupled} optimizer, with a learning rate of $6 \times 10^{-5}$, betas of (0.9, 0.999), and weight decay of 0.01. The models are initialized using the pretrained weights from ADE20K \cite{zhou2017scene}. We observe that it generally takes less than 20,000 iterations or \emph{less than five hours} to converge on 8 NVIDIA A100 GPUs. Note that we do not focus on the the architecture design in this work.

\textbf{Efficient Adaption to Unseen Models.} Generalizing to unseen domains is challenging for deep networks, yet it is necessary for practical usage. Therefore, we also explore how our pretrained artifacts detector could generalize to totally unseen generative models. In the experiment, we find that our pretrained model can detect a reasonable amount of perceptual artifacts in the unseen models. Furthermore, we find that fine-tuning our pretrained model with as few as ten images can quickly and effectively improve segmentation performance. The results are discussed in section \ref{subsec:unseen_methods}. Fine-tuning converges within 2,000 iterations, which would take \emph{less than 30 minutes}. Labeling 10 images from an unseen method would take \emph{approximately 10 minutes}, making this approach practical. Therefore, our pretrained model allows for fast adaptation to unseen models with roughly one hour of effort, making our work applicable and easily scalable in the future.

\subsection{Framing the Inpainting Context by Zoom-In} 
\label{subsec:artifacts_fix}

An important application of our artifacts detector is to automatically fix the perceptual artifacts region in the generated images. To construct this pipeline, a simple approach involves sequentially stacking the pretrained artifacts detector ($F$) and an inpainting model ($G$). During inference, we first feed the generated image ($I_g$) into the artifacts detector ($F$) to segment the artifacts region. Next, we consider the segmented artifacts mask with slight dilation as the inpainting mask for $I_g$ and process it through the inpainting model ($G$). The resulting artifacts-corrected image is denoted as $I_f$, and the overall pipeline can be expressed as $I_f = G(I_g, \phi(F(I_g)))$, where $\phi$ represents dilation.

Although naive inpainting can effectively fix many perceptual artifacts, we have observed systematic errors in recent diffusion-based inpainting models \cite{rombach2022high, ramesh2022hierarchical} when generating specific object details such as faces and hands. However, are these models genuinely incapable of generating such details? We conjecture that one potential cause of this error might be incorrect inpainting context. For instance, we have noticed that image generation generally has higher fidelity when human faces or hands are relatively prominent in the image. This could be due to two underlying reasons: 1) photographs or portraits of humans are typically large and centered in images, resulting in dataset bias; and 2) the loss on large object regions tends to be relatively greater than that of small objects, providing stronger feedback to the model during training.

\begin{figure}[!h]
    \centering
    \includegraphics[trim=0.0in 1.5in 4.9in 0in, clip,width=0.47\textwidth]{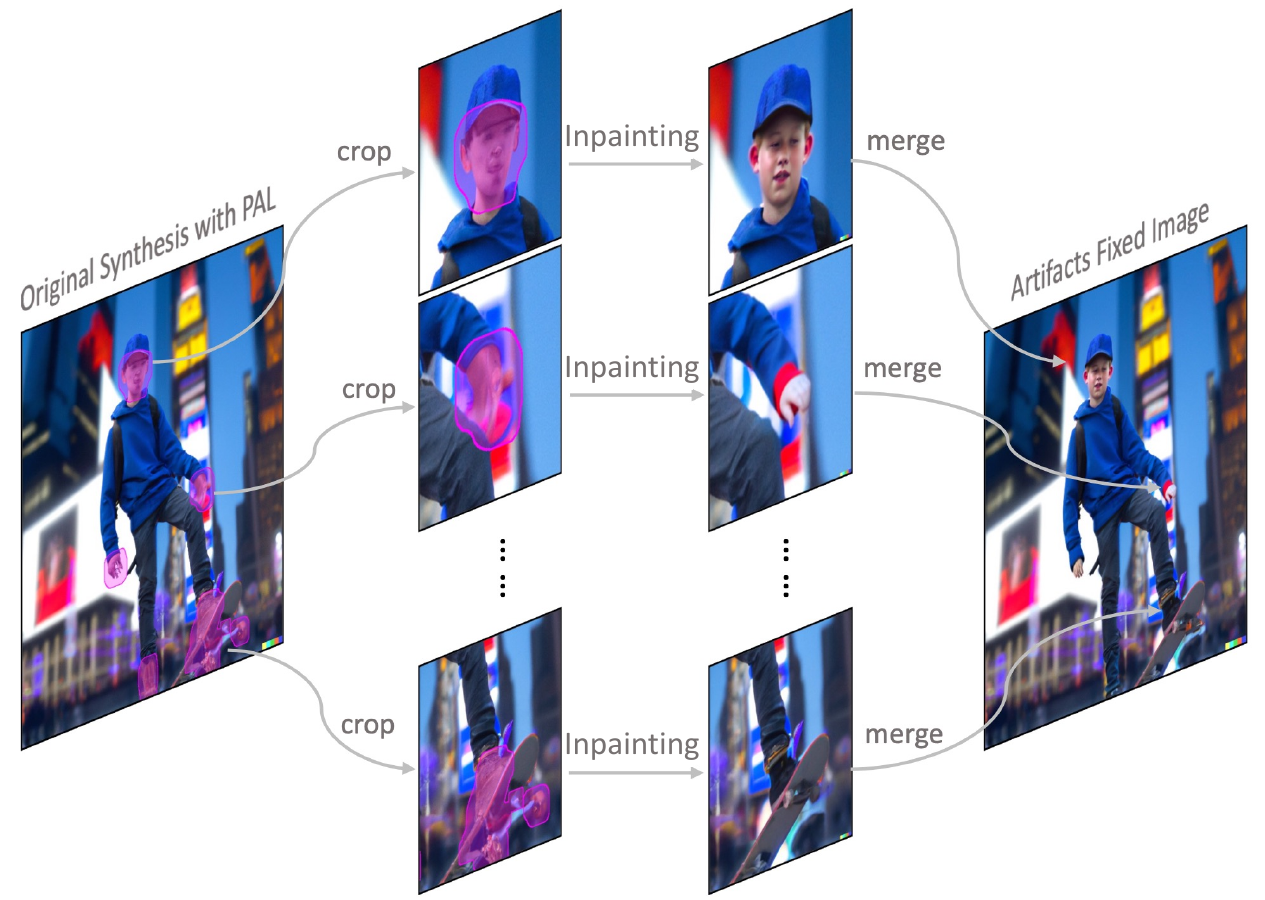}
    \caption{A zoom-in inpainting pipeline refines perceptual artifacts. Starting with the generated image that has predicted perceptual artifacts, we first crop around the artifact regions, using connected components as a guide. We then inpaint these artifact regions within each cropped area and ultimately composite them back into the full image. }
    \label{fig:zoom_in_pipeline}
\end{figure}

Motivated by this realization, we introduce a zoom-in inpainting pipeline that properly frames the input inpainting context before generating the output. As illustrated in Figure \ref{fig:zoom_in_pipeline}, we first conduct connected component analysis on the predicted artifacts segmentation mask, then crop around each component of artifacts, perform inpainting on these zoomed-in patches, and finally merge the inpainted patches back into the full image. We empirically set the patch size to be $50\%$ larger than the length of the longest axis of the artifacts mask within each connected component. Remarkably, this simple design significantly enhances artifacts refinement on object details \emph{without modifying the inpainting models}, as demonstrated in section \ref{sub_sec: inpainting_context}.

\section{Experiments}

\subsection{Performance on Diverse Image Synthesis Tasks}

Our main goal is to develop artifact detectors that can effectively perform on a wide range of generic image synthesis tasks and detect various types of artifacts. In order to achieve this, our first step is to gain a deeper understanding of how existing relevant methods would perform on this particular task. To begin with, we investigate the CNNgenerates \cite{wang2022pretraining} classifier, which is specifically designed to distinguish between generated and real images in the context of generic deep generative models. To visualize the gradient activation of the model, we employ Grad-CAM \cite{selvaraju2017grad}, which could reveal regions in the image that are deemed as "fake" by the network. Nonetheless, our findings demonstrate a significant disparity between the model's interpretation and human perception of what constitutes "fake" or artifacts, as presented in the $1^{st}$ row of Table \ref{tab:unified_vs_specialist}. In addition, we leverage the Patch Forensics model \cite{chai2020makes} to calculate the "fake" regions based on the patch-based classifier. The results demonstrate that the model's prediction also significantly deviates from human perception, as shown in $2^{nd}$ row of Table \ref{tab:unified_vs_specialist}. Related to our work, PAL4Inpaint \cite{zhang2022perceptual} focuses on developing a perceptual artifacts localization method for the inpainting task. However, we observe that a model trained exclusively on inpainting images struggles to generalize to other tasks, as shown in $3^{rd}$ of Table \ref{tab:unified_vs_specialist}.

The aforementioned observations from previous studies underscore the compelling need for a diverse dataset encompassing multiple tasks and domains to train a generalized artifacts detector. In light of this, we collect a fine-grained labeled dataset spanning ten image synthesis tasks. Subsequently, we train specialized models for each task, as well as a single unified model for all tasks, as shown in the last two rows of Table \ref{tab:unified_vs_specialist}. The unified model confers a memory-efficient advantage from the deployment perspective and performs comparably to the specialist models, with the exception of portrait shadow removal (PSR) task. We believe that this discrepancy may be attributed to the dissimilarity between the artifacts in PSR task and those in other tasks. Nevertheless, our specialist models and unified model both demonstrate significant superior performance in contrast to existing methods.

\begin{figure*}[!b]
    \centering
    \includegraphics[trim=0.0in 1.3in 4.5in 0in, clip,width=\textwidth]{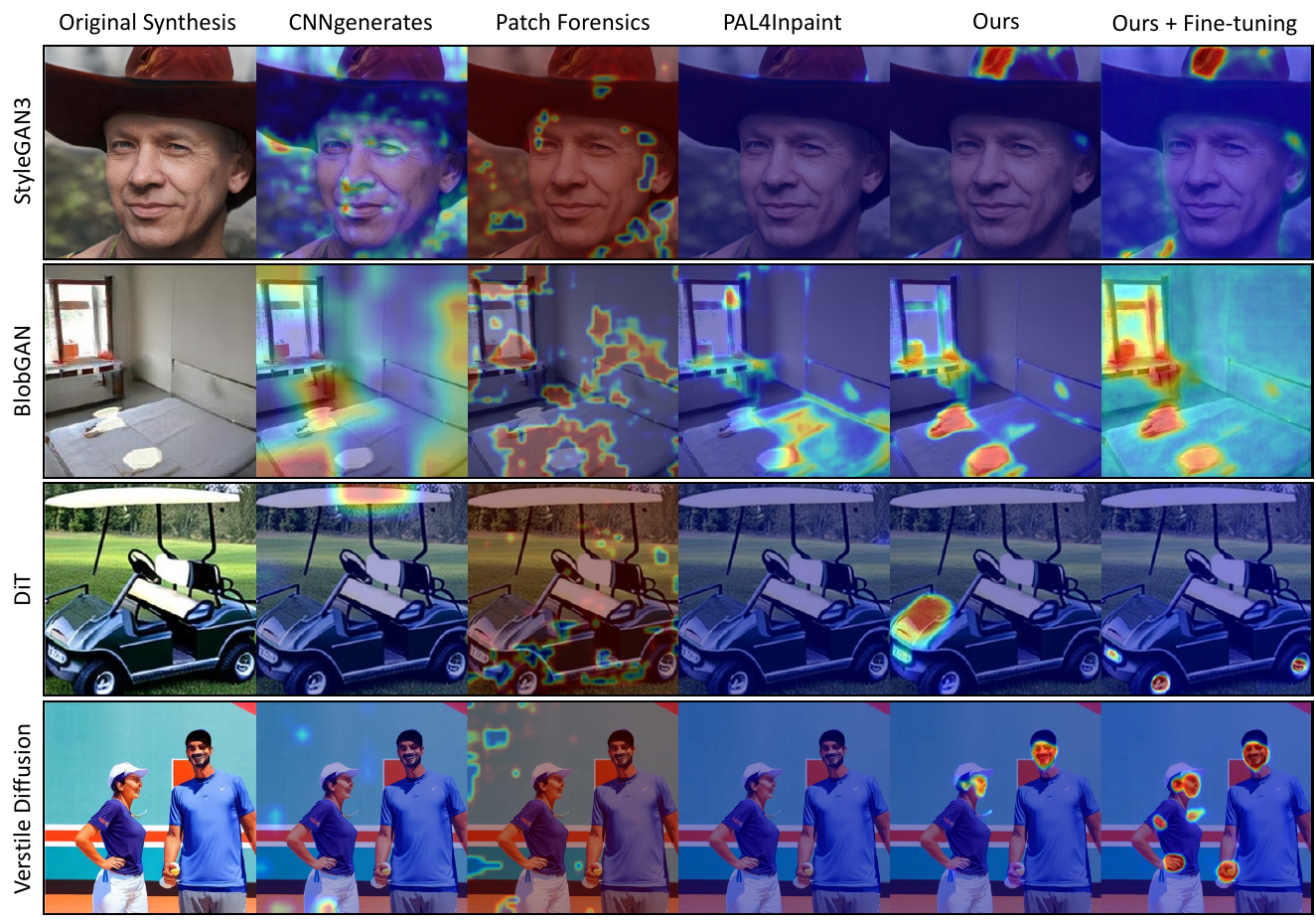}
    \caption{Qualitative comparison of artifacts localization on several unseen methods. We use Grad-CAM \cite{selvaraju2017grad} to visualize the gradient maps of CNNgenerates \cite{wang2020cnn}, and use the pretrained checkpoints from Patch Forensics \cite{chai2020makes} and PAL4Inpaint \cite{zhang2022perceptual} to directly compute the heatmap. The results demonstrate that our approaches exhibit a much stronger correlation with human judgement in detecting perceptual artifacts. \textbf{Please zoom in the first column to check the perceptual artifacts.} }
    \label{fig:unseen_comparison}
\end{figure*}

\begin{figure*}[!t]
    \centering
    \includegraphics[trim=0in 0.4in 2.9in 0in, clip,width=\textwidth]{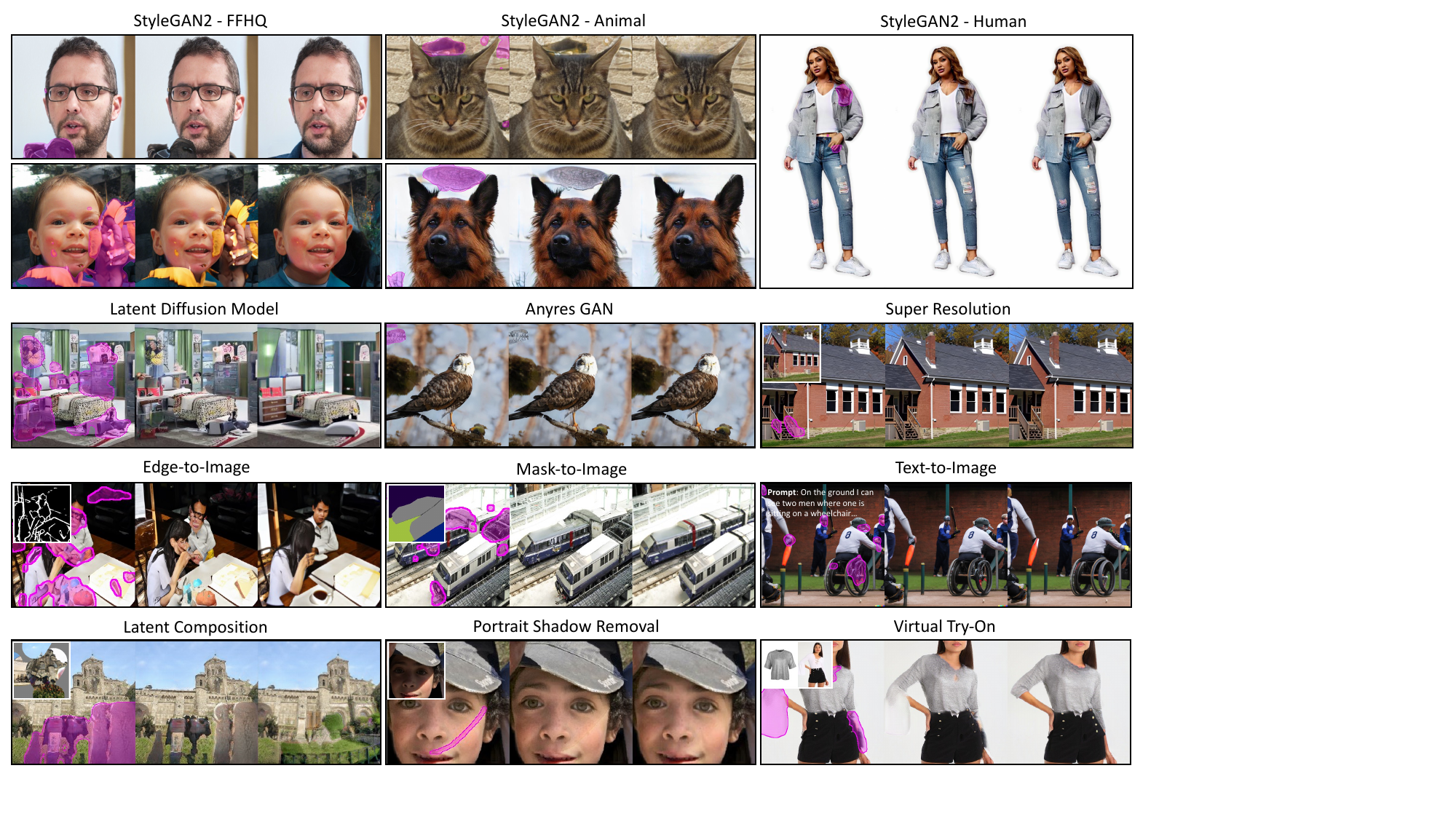}
    \caption{Qualitative results of automatic artifacts fixing on ten image synthesis tasks. In each grid, the {\color{RubineRed}left is the overlaid artifacts segmentation}, the {\color{ForestGreen}middle is the original generated image}, and the {\color{NavyBlue}right is the refined image}. }
    \label{fig:ten_task_fix}
\end{figure*}

\begin{figure*}[!t]
    \centering
    \includegraphics[trim=0.1in 5.0in 1.0in 0in, clip,width=\textwidth]{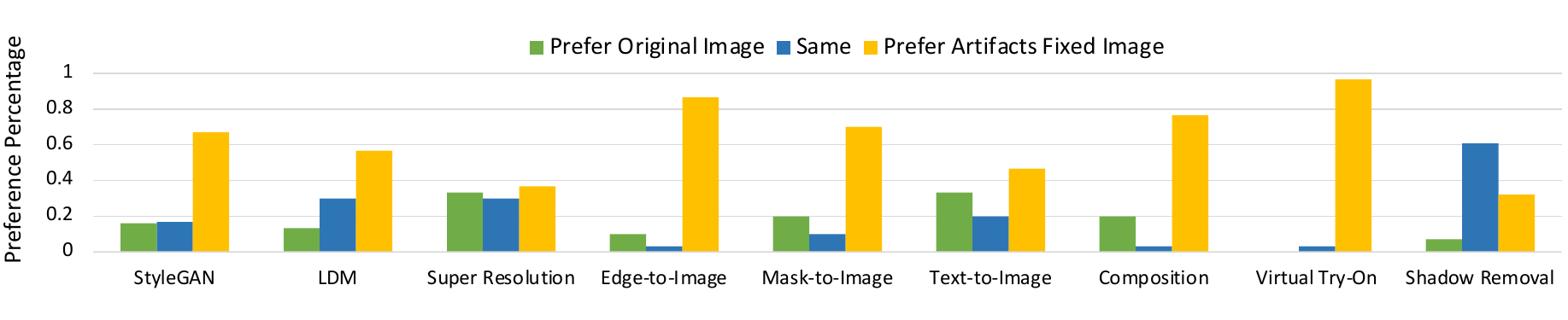}
    \caption{User study to evaluate whether the curated images are better, similar, or worse than the original generated images for diverse synthesis tasks. For each task, we sample 30 images and ask at least five users to vote for each image.  }
    \label{fig:fix_user_study}
\end{figure*}

\subsection{Performance on Unseen Methods}
\label{subsec:unseen_methods}

Given the rapidly evolving landscape of generative image models, with novel models emerging on a monthly basis, an ideal artifacts detector should be able to effectively function or swiftly adapt to these untested methods. To evaluate the performance on the unseen methods, we collected additional 500 generated images with labels from two previously unseen GAN-based models and three diffusion-based models. Note that our dataset includes images from StyleGAN2 and Stable Diffusion v1.4, but excludes any images from StyleGAN3 and Stable Diffusion v2.0 (SD2). The other three models are BlobGAN \cite{epstein2022blobgan}, Verstile Diffusion \cite{xu2022versatile} and Diffusion Transformer (DiT) \cite{peebles2022scalable}, which do not have any counterparts in our training dataset. 

\begin{table}[!h]
    \begin{center}
     \resizebox{0.48\textwidth}{!}{
      \begin{tabular}{l|c|c|c|c|c}
        \toprule 
        \textbf{Methods} & \ StyleGAN3 \cite{karras2021alias} & \ BlobGAN \cite{epstein2022blobgan} & \ SD2 \cite{rombach2022high} & \ VD \cite{xu2022versatile} & \ DiT \cite{peebles2022scalable}  \\
        \midrule 
        CNNgen \cite{wang2020cnn} + \cite{selvaraju2017grad} & 2.30 & 3.67 & 0.12 & 0.57 & 1.42  \\
        Patch Forensics \cite{chai2020makes} & 11.43 & 5.96 & 3.08 & 2.97 & 3.18 \\
        PAL4Inpaint \cite{zhang2022perceptual} & 5.18 & 13.0 & 0.85 & 0.63 & 1.15 \\
        \midrule
        Ours & \textbf{46.45} & 25.39 & 6.75 & 5.92 & 16.46 \\
        Ours w/ Fine-tuning & 40.81 & \textbf{33.33} & \textbf{11.04} & \textbf{22.18} & \textbf{31.76} \\
        \bottomrule
      \end{tabular}
      }
      \caption{Quantitative mIoU ($\uparrow$) evaluation of the binary artifacts segmentation on five unseen models. We use the following brevity to indicate the tasks: CNNgen $\rightarrow$ CNNgenerates \cite{wang2020cnn}, SD2 $\rightarrow$ Stable Diffusion v2.0 \cite{rombach2022high}, VD $\rightarrow$ Verstile Diffusion \cite{xu2022versatile}, DiT $\rightarrow$ Diffusion Transformer \cite{peebles2022scalable}.  }
      \label{tab:unseen_comparison}
    \end{center}
\end{table}



We first visualize what the previous and our methods detect as visual artifacts in the generated images from unseen methods. To this ends, we compute the raw heatmap outputs and compare them in Figure \ref{fig:unseen_comparison}. For quantitative comparison in Table \ref{tab:unseen_comparison}, the previous methods exhibit poor performance in segmenting the perceptual artifacts in these unseen images. In contrast, our unified model trained on the newly proposed dataset demonstrates reasonable generalization ability to these unseen models. Furthermore, fine-tuning our model with a minimum of 10 examples results in effective performance improvement.

\begin{figure*}[!t]
    \centering
    \includegraphics[trim=0in 5.0in 2.0in 0in, clip,width=\textwidth]{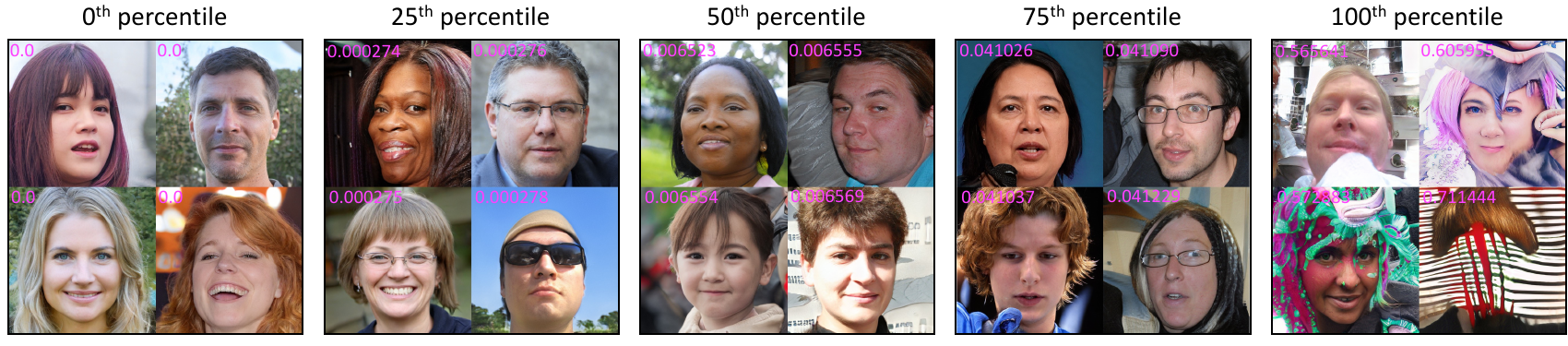}
    \caption{Using Perceptual Artifacts Ratio (PAR) as a metric to rank 4,000 synthesized face image by StyleGAN2. We show samples at different percentile from the PAR ranking, where the actual PAR score is written in pink at the top left corner of each image. }
    \label{fig:rank_ffhq}
\end{figure*}

\begin{table*}[!t]
\begin{minipage}[b]{0.6\linewidth}
\centering
\includegraphics[trim=0in 4.9in 5.5in 0in, clip,width=\textwidth]{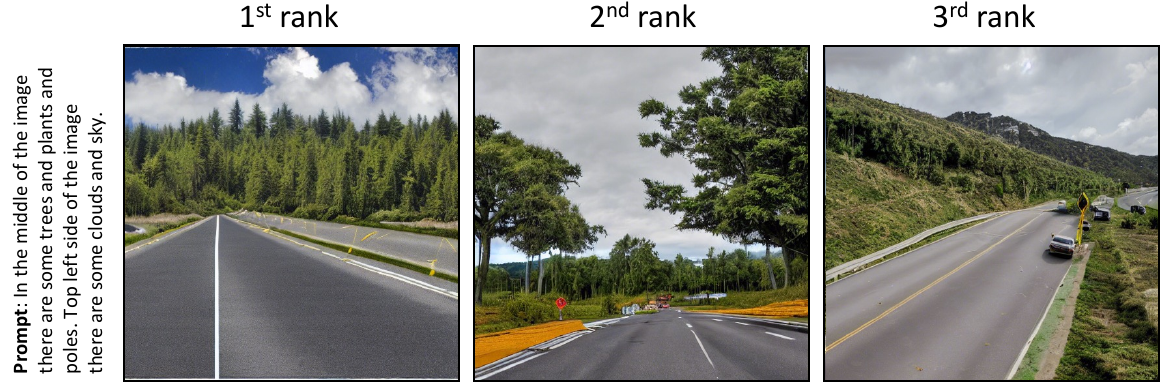}
\captionof{figure}{Using Perceptual Artifacts Ratio (PAR) to rank the multimodal outputs of text-to-image synthesis by Stable Diffusion. }
\label{fig:rank_sd}
\end{minipage}
\begin{minipage}[b]{0.38\linewidth}
\centering
    \resizebox{1.0\textwidth}{!}{
      \begin{tabular}{l|c|c}
        \toprule 
        \textbf{Tasks} \hspace{0pt} & \hspace{0pt} \ StyleGAN2 \cite{karras2020analyzing} \hspace{0pt} & \hspace{0pt} \ Stable Diffusion \cite{rombach2022high}  \hspace{0pt}  \\
        \midrule 
        Random Chance & 50.0\% & 50.0\% \\
        \midrule
        HyperIQA \cite{su2020blindly} & 58.51\% \color{ForestGreen}(+8.51\%) & 43.30\% \color{BrickRed}(-6.70\%) \\
        MUSIQ \cite{ke2021musiq} & 61.63\% \color{ForestGreen}(+11.63\%) & 48.45\% \color{BrickRed}(-1.55\%)\\
        PAR (Ours) & 74.47\% \color{ForestGreen}(+24.47\%) & 63.92\% \color{ForestGreen}(+13.92\%) \\
        \bottomrule
      \end{tabular}
      }
    \caption{A quantitative study of user agreement with metrics to rank the visual quality between pairs of images in unconditional StyleGAN2 \cite{karras2020analyzing} sampling and multimodal text-to-image synthesis with Stable Diffusion \cite{rombach2022high}. We show comparison with two state-of-the-arts non-referenced IQA methods \cite{rombach2022high, ke2021musiq}. }
    \label{tab:user_study_par_rank}
\end{minipage}\hfill
\end{table*}

\subsection{Automatically Fixing Artifacts}

An essential downstream application of perceptual artifacts localization is the automatic correction or refinement of artifacts in generated images, as discussed in Section \ref{subsec:artifacts_fix}. In this section, we provide both qualitative (Figure \ref{fig:ten_task_fix}) and quantitative demonstrations of how perceptual artifacts segmentation is capable of effectively correcting a significant portion of the perceptual artifacts in diverse synthesis tasks. 


To quantitatively measure the artifacts fixing performance, we conduct a user study to assess whether the artifacts corrected images are better, similar, or worse than the original generated images. For each task, we randomly select approximately 30 images and solicit feedback from at least five Amazon Turk workers to vote on each pair of images. As depicted in Figure \ref{fig:fix_user_study}, the user preferences demonstrate that our artifacts correction pipeline significantly enhances the visual quality for most of the tasks, and rarely degrades the quality of the generated images. Although we believe that the user study provides a more accurate reflection of perceptual improvement, we also calculate the FID-CLIP \cite{kynkaanniemi2022role} of 5,000 Stable Diffusion (SD) \cite{rombach2022high} text2image images, which improves from 15.98 to 13.28 (+16.9\%) after our refinement with SD inpainting \cite{rombach2022high}.

\subsection{Perceptual Artifacts as A Quality Metric}

Evaluating the quality of generated images remains an ongoing area of research. Among various types of image quality assessment (IQA), no-reference IQA is the most challenging, as there are no reference ground truth images for comparison. In this study, we demonstrate that our artifacts detector can be leveraged to compute a no-reference IQA metric referred to as Perceptual Artifacts Ratio (PAR), which is calculated as the ratio of the perceptual artifacts region to the entire image area. Essentially, a larger PAR value indicates more perceptual artifacts in the image and, consequently, lower visual quality.

We demonstrate the application of the PAR metric to rank unconditional and multimodal image samples. In Figure \ref{fig:rank_ffhq}, we showcase how PAR can rank thousands StyleGAN2 face images, where smaller PAR values indicate fewer artifacts and, thus, better visual quality. We extract four images at percentiles of $0^{th}$, $25^{th}$, $50^{th}$, $75^{th}$, and $100^{th}$, where $0^{th}$ and $100^{th}$ percentiles correspond to the least and largest PAR, respectively. Our results indicate that the visual quality gradually deteriorates with increasing percentiles, which is consistent with human perception. Additionally, we demonstrate that the PAR score can help rank the visual quality of multimodal text-to-image outputs generated by Stable Diffusion \cite{rombach2022high}, as illustrated in Figure \ref{fig:rank_sd}.

Furthermore, we conduct a user study to evaluate user preference agreement with the no-reference IQA metrics for ranking around 100 pairs of images for both StyleGAN2 \cite{karras2020analyzing} and Stable Diffusion \cite{ramesh2022hierarchical}. The results presented in Table \ref{tab:user_study_par_rank} indicate that our PAR metric outperforms two state-of-the-art methods in terms of ranking image quality. In practical use cases, the PAR score can facilitate automatic ranking or filtering of a large batch of candidate images.

\subsection{Effect of Zoomed-in Inpainting Context}
\label{sub_sec: inpainting_context}

In section \ref{subsec:artifacts_fix}, we discussed how diffusion-based inpainting models tend to produce better outputs for object details when zoomed-in context is provided. In this section, we aim to quantitatively evaluate the impact of zoom-in inpainting on artifacts refinement performance. It is widely known that diffusion models, such as Stable Diffusion \cite{rombach2022high} or DALL-E 2 \cite{ramesh2022hierarchical}, often struggle with generating realistic human faces and hands. Therefore, we investigate how zoom-in inpainting can aid in refining these challenging cases.

\begin{figure}[!h]
    \centering
    \includegraphics[trim=0.0in 3.6in 4.3in 0in, clip,width=0.47\textwidth]{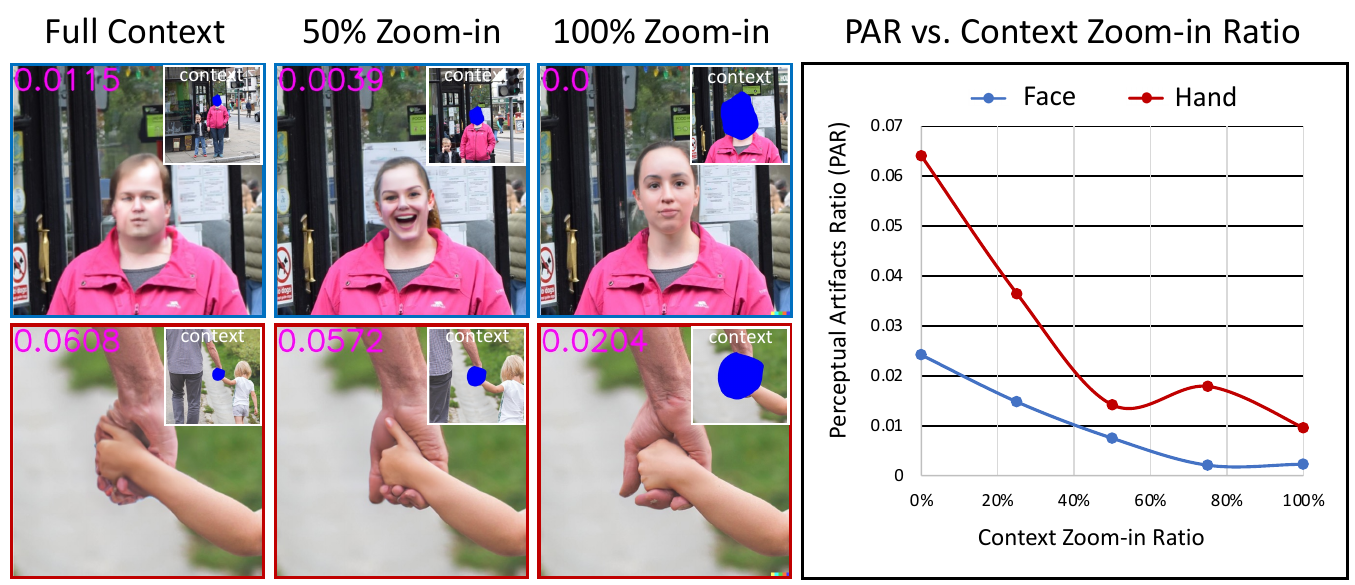}
    \caption{The relationship between perceptual artifacts ratio (PAR) and context zoom-in. In each image of left examples, top right is the input context for inpainting. }
    \label{fig:dalle_bias}
\end{figure}

As shown in Figure \ref{fig:dalle_bias}, the two examples on the left illustrate that gradually zooming-in the input context leads to the generation of more realistic face and hand pixels. To further support our findings, we employed the PAR score to quantify the relationship between zoomed-in scale and image quality. Our results demonstrate that providing gradually zoomed-in context consistently enhances the quality of generated face and hand pixels over a set of 100 images, as shown on the right of Figure \ref{fig:dalle_bias}. Hence, we can observe that our simple zoom-in inpainting pipeline effectively addresses the artifacts present in these object details.

\subsection{Abnormal Detection on Real Images}

\begin{figure}[!h]
    \centering
    \includegraphics[trim=0.0in 0.7in 0.5in 0in, clip,width=0.48\textwidth]{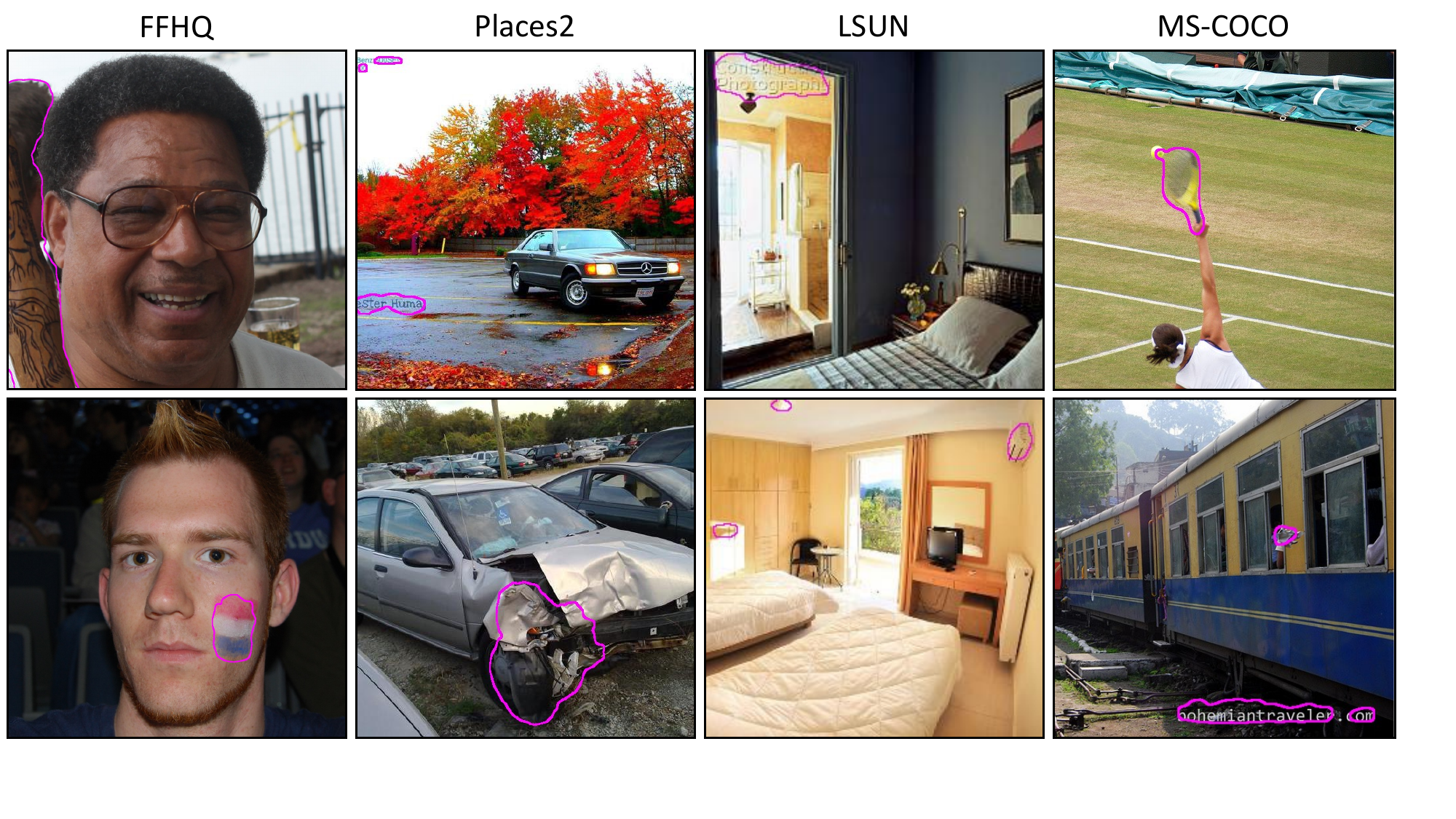}
    \caption{Inference on real natural images, where the predicted abnormal regions are indicated by the pink contour. }
    \label{fig:real_images}
\end{figure}

Since our artifacts detector demonstrates reasonable perceptual artifacts prediction ability on photo-realistic generated images, we are curious if the model would detect anything unusual in real images. As expected, the majority of the real images, which do not contain any artificially generated content, receive no prediction from the artifacts detector. Interestingly, for a small portion of images that receive some prediction, we observe that the predicted perceptual artifacts tend to be on abnormal objects, distractors, or blurry/fuzzy regions, as shown in several examples in Figure \ref{fig:real_images}. For instance, in an FFHQ \cite{karras2019style} face image, the artifacts detector finds an "abnormal" object that appears to be a tattooed arm. For other in-the-wild images sampled from commonly used datasets \cite{zhou2017places, lin2014microsoft, yu2015lsun}, the artfacts detector model also detects artifacts such as watermark text, fuzzy or distractor bedroom corners, and tennis rackets with motion blur, as shown in the right three images of Figure \ref{fig:real_images}.

\section{Conclusion}

In this paper, we present a comprehensive empirical study on perceptual artifacts localization for image synthesis tasks. Firstly, we collected a high-quality dataset comprising 10,168 images with per-pixel artifact labels. Subsequently, we trained a segmentation model to accurately locate the artifacts for ten diverse synthesis tasks, and demonstrated that our pre-trained model can efficiently adapt to unseen methods. Utilizing our learned artifact detector, we explore three downstream applications: 1) automatically refining artifacts in the generated images; 2) evaluating image quality without reference; and 3) detecting abnormal objects in natural images. To address the issue of diffusion-based inpainting generating incorrect content in object details such as faces and hands, we propose a simple zoom-in inpainting pipeline that effectively mitigates this problem. 

\textbf{Future Directions.} We view our dataset and benchmark models as foundational for future research in this domain. We identify three primary avenues for advancement: 1) crafting specialized architecture or loss functions for enhanced perceptual artifact segmentation; 2) creating task-specific modules to achieve more granular refinement for each task, moving beyond mere inpainting;  and 3) broadening the dataset to encompass varied individual preferences concerning perceptual artifacts in generated images.

\textbf{Practical Impact.} To the best of our knowledge, our research on the localization of perceptual artifacts in generic image synthesis is pioneering. Manually retouching these perceptual artifacts can be both tedious and vexing, especially for professionals. Our automated approach stands to greatly enhance productivity and alleviate this significant burden. Furthermore, our no-reference perceptual artifacts metric can assist users in sifting through and selecting quality content from multimodal generation candidates, such as text-to-image applications \cite{rombach2022high, ramesh2022hierarchical}, potentially boosting user satisfaction. In summary, we hope that our contributions might offer valuable insights and tools for the evolution of practical image editing software in the coming years.

{\small
\bibliographystyle{ieee_fullname}
\bibliography{egbib}
}

\newpage
\title{\large Supplementary Materials}  

\maketitle
\thispagestyle{empty}
\appendix

\section{Details on Data Labeling}


We have discussed about the overall data labeling and statistics in the main paper. Here, we added more details regarding the data labeling. 

\textbf{Labeling Criterion} Labeling perceptual artifacts is a highly subjective task, and therefore different workers may have varying opinions on which regions should be considered as 'artifacts'. We instruct the workers to keep a specific criterion in mind while labeling, which is to imagine that we have a perfect artifact-fixing model that can correct any marked region. Hence, if a worker believes that any region in the image can be enhanced or refined, they should mark those regions accordingly.

\textbf{User Interface} We use a labeling interface similar to the one used in \cite{zhang2022perceptual}, where we duplicate the generated image and stitch the copies side-by-side. During labeling, workers can identify perceptual artifacts on the right side of the image and refer to the 'unmarked' image on the left side as a reference.

\begin{figure*}[!h]
 \vspace{-10 pt}
    \centering
    \includegraphics[trim=0in 0.7in 0.6in 0in, clip,width=\textwidth]{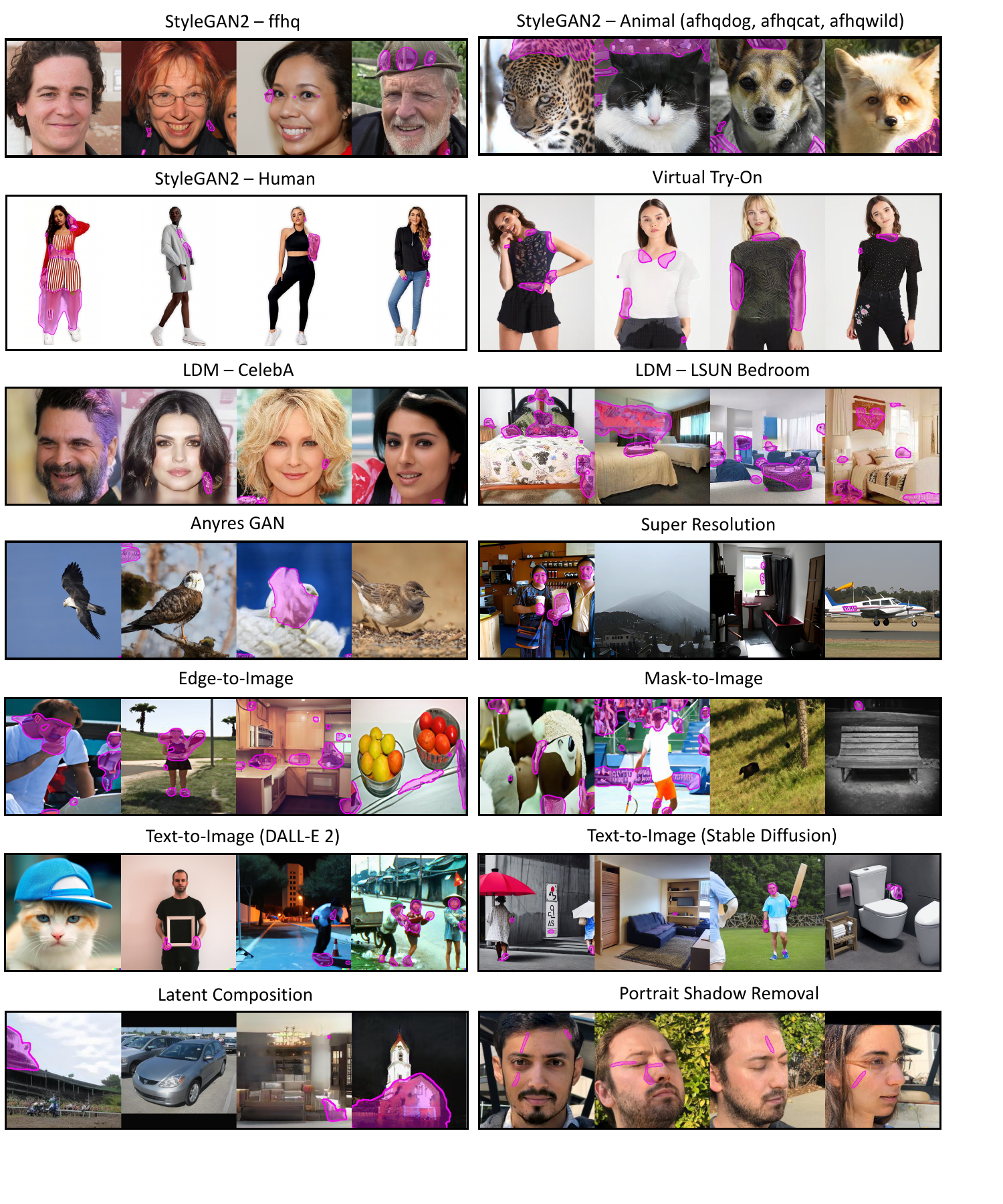}
    \vspace{-5 pt}
    \caption{A sampled visualization of our labeled perceptual artifacts dataset in diverse synthesis tasks and domains. Note that if there is no mask in the image, it indicates that workers do not think there are any artifacts in the generated image. }
    \label{fig:supp_dataset}
    \vspace{-5 pt}
\end{figure*}

\textbf{Visualization of Labels} We show more visualization of the perceptual artifacts labels in Fig \ref{fig:supp_dataset}.

\section{Implementation Details}

\textbf{Training Details of PAL Models} We implement our PAL model using the Swin-L as the backbone, UperNet as the head with a loss weight of 1.0, and a FCN auxiliary head with a loss weight of 0.4. During training, we use random crop with max cutout ratio of 0.75, and random flip with a probability of 0.5. Our code implementation is based on MMSegmentation \footnote[1]{mmsegmentation: \href{https://github.com/open-mmlab/mmsegmentation}{https://github.com/open-mmlab/mmsegmentation}}. 

\textbf{Code Resources for Data Generation} We use the official github repos to generate the images for each synthesis tasks. The StyleGAN 2 images from domain ffhq, afhqdog, afhqcat, and afhqwild are generated using the official NVIDIA StyleGAN repo \footnote[2]{stylegan: \href{https://github.com/NVlabs/stylegan3}{https://github.com/NVlabs/stylegan3}}. For StyleGAN 2 human images, we use the StyleHuman repo \footnote[3]{StyleGAN-Human: \href{https://github.com/stylegan-human/StyleGAN-Human}{https://github.com/stylegan-human/StyleGAN-Human}}. For unconditional generation with Latent-Diffusion Model (LDM), we use the code from \footnote[4]{latent-diffusion: \href{https://github.com/CompVis/latent-diffusion}{https://github.com/CompVis/latent-diffusion}}. We generate Anyres GAN images using \footnote[5]{anyres-gan: \href{https://github.com/chail/anyres-gan}{https://github.com/chail/anyres-gan}}, and super resolution with Real-ESRGAN \footnote[6]{Real-ESRGAN: \href{https://github.com/xinntao/Real-ESRGAN}{https://github.com/xinntao/Real-ESRGAN}}. For Edge-to-Image and Mask-to-Image, we use the same diffusion-based model PITI \footnote[7]{PITI: \href{https://github.com/PITI-Synthesis/PITI}{https://github.com/PITI-Synthesis/PITI}} but different checkpoints. For DALL-E 2 text-to-image synthesis and inpainting, we use the OpenAI API \footnote[8]{dalle-api: \href{https://openai.com/api/}{https://openai.com/api/}}. For Stable Diffusion, we use the v1.4 checkpoint from \footnote[9]{stable-diffusion-v1.4: \href{https://github.com/CompVis/stable-diffusion}{https://github.com/CompVis/stable-diffusion}} for text-to-image synthesis, and v1.5 checkpoint from \footnote[10]{stable-diffusion-v1.5: \href{https://github.com/runwayml/stable-diffusion}{https://github.com/runwayml/stable-diffusion}} for inpainting. We use official latent composition repo \footnote[11]{latent-composition: \href{https://github.com/chail/latent-composition}{https://github.com/chail/latent-composition}} for image composition synthesis. Finally, we directly use the synthesized images from repo \footnote[12]{c-vton: \href{https://github.com/benquick123/C-VTON}{https://github.com/benquick123/C-VTON}} for virtual try-on task and repo \footnote[13]{portrait-shadow-manipulation: \href{https://github.com/google/portrait-shadow-manipulation}{https://github.com/google/portrait-shadow-manipulation}} for portrait shadow removal. For other inpainting models used in artifacts fixing pipeline, we use official LaMa github repo \footnote[14]{lama: \href{https://github.com/saic-mdal/lama}{https://github.com/saic-mdal/lama}}, and official CoMod-GAN github repo \footnote[15]{co-mod-gan: \href{https://github.com/zsyzzsoft/co-mod-gan}{https://github.com/zsyzzsoft/co-mod-gan}}.

\textbf{Prompt for Text-based Inpainting} Text-based diffusion inpainting requires additional text prompt besides the image and mask inputs. In this section, we discuss how we decide the fixed text prompts for each type of generated images. Generally, we use "a person's face" as the text prompt for all facial images generated by StyleGAN2 \cite{karras2020analyzing} and LDM \cite{rombach2022high}, and use "a person" for all human images in StyleGAN2 human and virtual try-on task. For LDM LSUN bedroom images, we just use "bedroom" as the text prompt. For the rest of in-the-wild images, we use "photograph of a beautiful empty scene, highest quality settings" as the fixed text prompt, which is the default option used in Stable Diffusion inpainting.

\textbf{Selecting Multimodal Outputs} As text-based inpainting models, i.e. DALL-E 2 \cite{ramesh2022hierarchical}, have multimodal outputs, we select the final output image based on the Perceptual Artifacts Ratio (PAR), which has some correlation with human judgement as described in section 5 of the main paper. Specifically, suppose we have $N$ multimodal outputs, we denote the candidate images as $I_{i}$, where $i = 1, ..., N$. We compute the PAR scores for each image, which is denoted as $PAR(I_i)$. The finally selected output image is determined by $\operatorname*{argmin}_i PAR(I_i)$.

\section{Statistical Analysis of User Study}

We conduct user studies to evaluate whether the artifacts fixed images are better, same, or worse the original generated images. We perform statistical hypothesis testing using a null hypothesis that the mean of preferences is zero, where the preference is -1 if the original image was preferred, 0 if no preference, and +1 if the artifacts-fixed image was preferred. We use a one sample permutation t test with $10^6$ permutations. If we combine all user votes into a single list, the null hypothesis is rejected with $p=0$. If we run a test per task, using Holm-Bonferroni correction and a familywise error rate of 0.05, we find the null hypothesis is rejected for every task except super-res, text-to-image, and shadow removal. This indicates that for 6 out 10 tasks and for the combination of all user votes across tasks, there is a significant preference, which per our data is the artifacts fixed image.

\section{More Qualitative Results}

In this section, we show more visualization results. 

\subsection{PAL and Artifacts Fixed Results}

\begin{figure*}[!t]
    \centering
    \includegraphics[trim=0in 1.9in 4.3in 0in, clip,width=\textwidth]{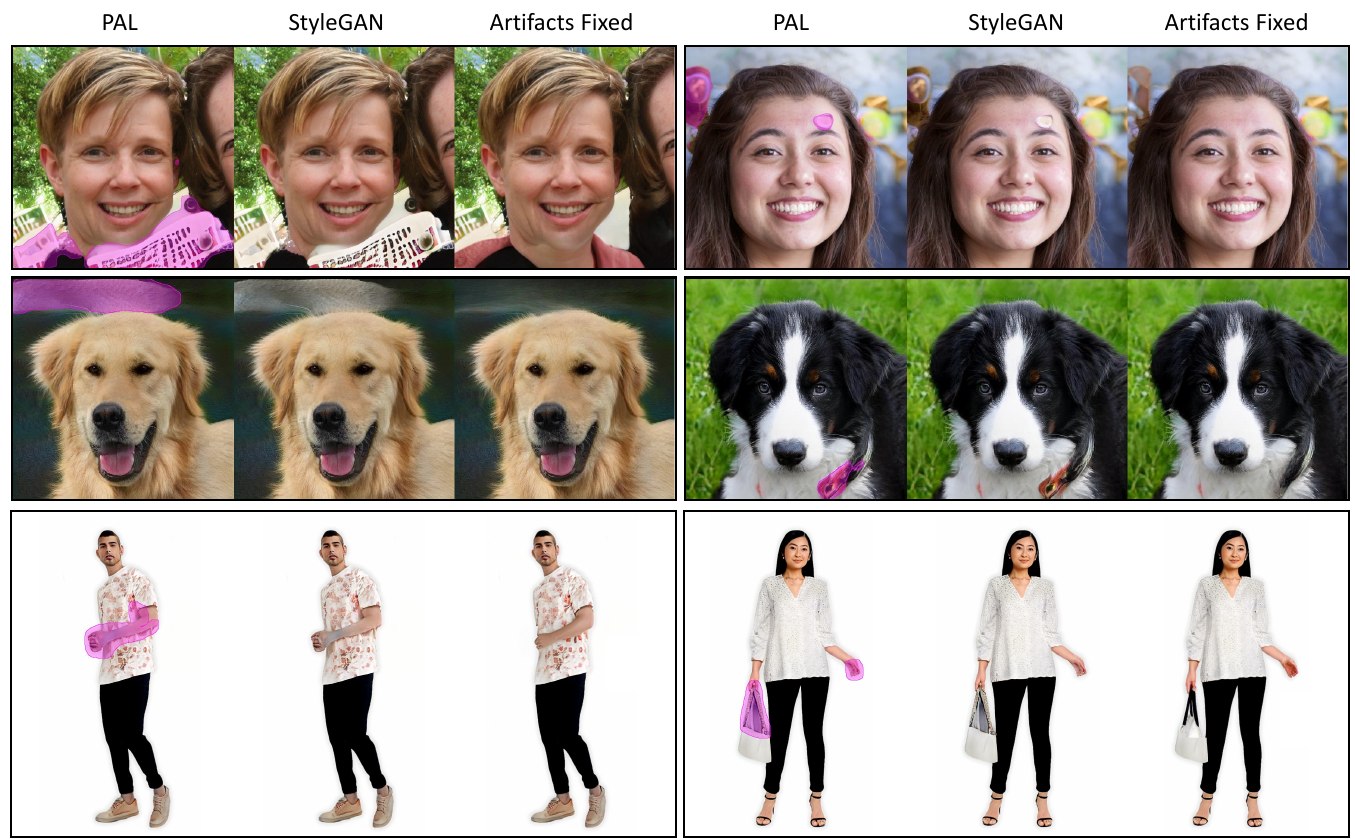}
    \vspace{-20 pt}
    \caption{More qualitative results for perceptual artifacts localization (PAL) prediction and the artifacts fixed images for StyleGAN \cite{karras2020analyzing}. \textbf{Left}: original generated image with PAL prediction. \textbf{middle}: original generated image. \textbf{right}: artifacts fixed/refined generated image. }
    \label{fig:supp_stylegan_fix}
    \vspace{-10 pt}
\end{figure*}

\begin{figure*}[!t]
    \centering
    \includegraphics[trim=0in 2.6in 4.3in 0in, clip,width=\textwidth]{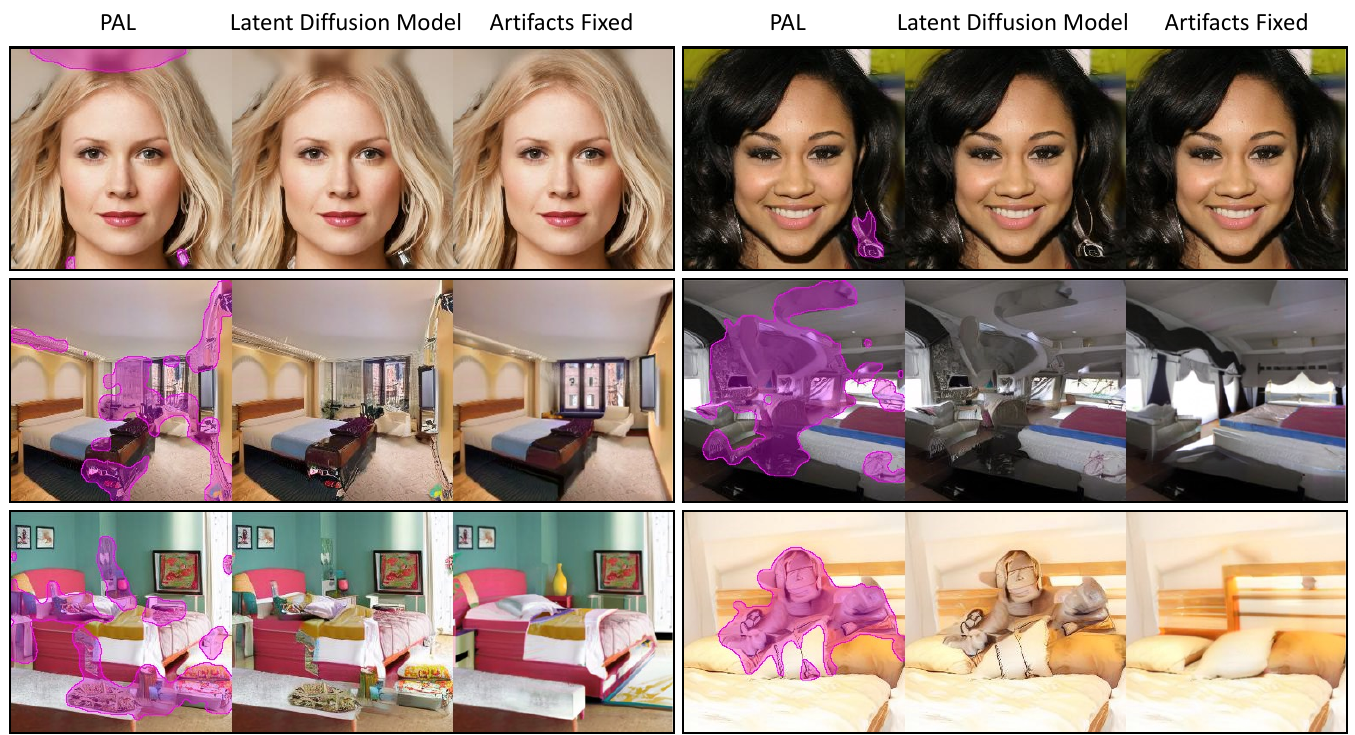}
    \vspace{-20 pt}
    \caption{More qualitative results for perceptual artifacts localization (PAL) prediction and the artifacts fixed images for Latent Diffusion Model \cite{rombach2022high}. \textbf{Left}: original generated image with PAL prediction. \textbf{middle}: original generated image. \textbf{right}: artifacts fixed/refined generated image.}
    \label{fig:supp_ldm_fix}
    \vspace{-5 pt}
\end{figure*}

\begin{figure*}[!h]
 \vspace{-10 pt}
    \centering
    \includegraphics[trim=0in 4.1in 4.3in 0in, clip,width=\textwidth]{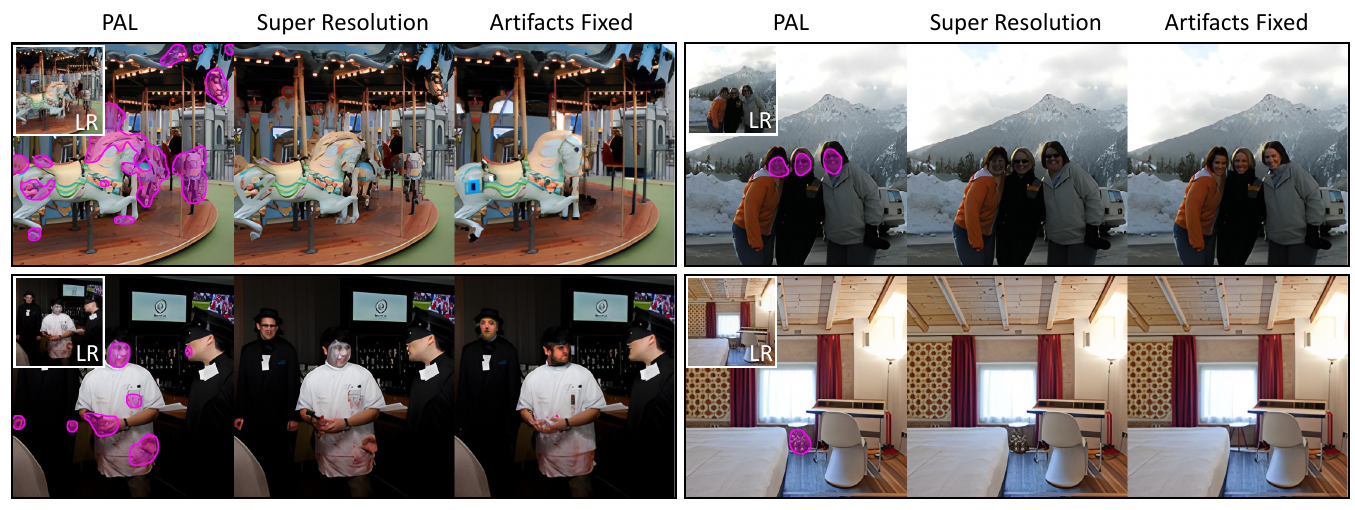}
    \vspace{-20 pt}
    \caption{More qualitative results for perceptual artifacts localization (PAL) prediction and the artifacts fixed images for super resolution with Real-ESRGAN \cite{wang2021real}.}
    \label{fig:supp_sr_fix}
    \vspace{-5 pt}
\end{figure*}

\begin{figure*}[!h]
 \vspace{-10 pt}
    \centering
    \includegraphics[trim=0in 4.1in 4.3in 0in, clip,width=\textwidth]{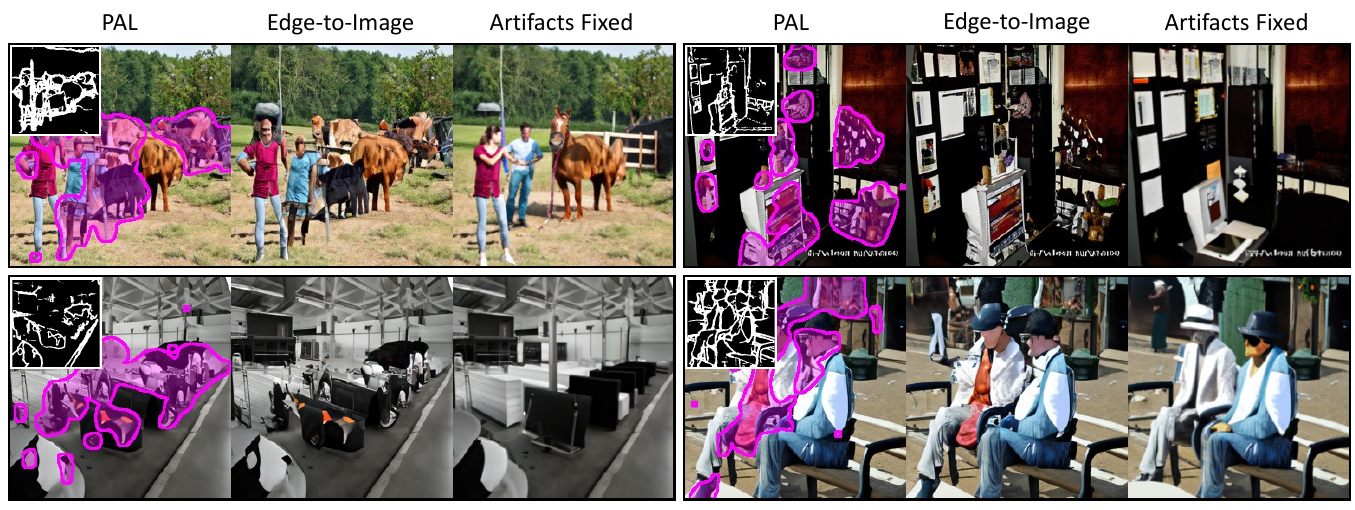}
    \vspace{-20 pt}
    \caption{More qualitative results for perceptual artifacts localization (PAL) prediction and the artifacts fixed images for Edge-to-Image translation with PITI \cite{wang2022pretraining}.}
    \label{fig:supp_edge2image_fix}
    \vspace{-5 pt}
\end{figure*}

\begin{figure*}[!h]
 \vspace{-10 pt}
    \centering
    \includegraphics[trim=0in 4.1in 4.3in 0in, clip,width=\textwidth]{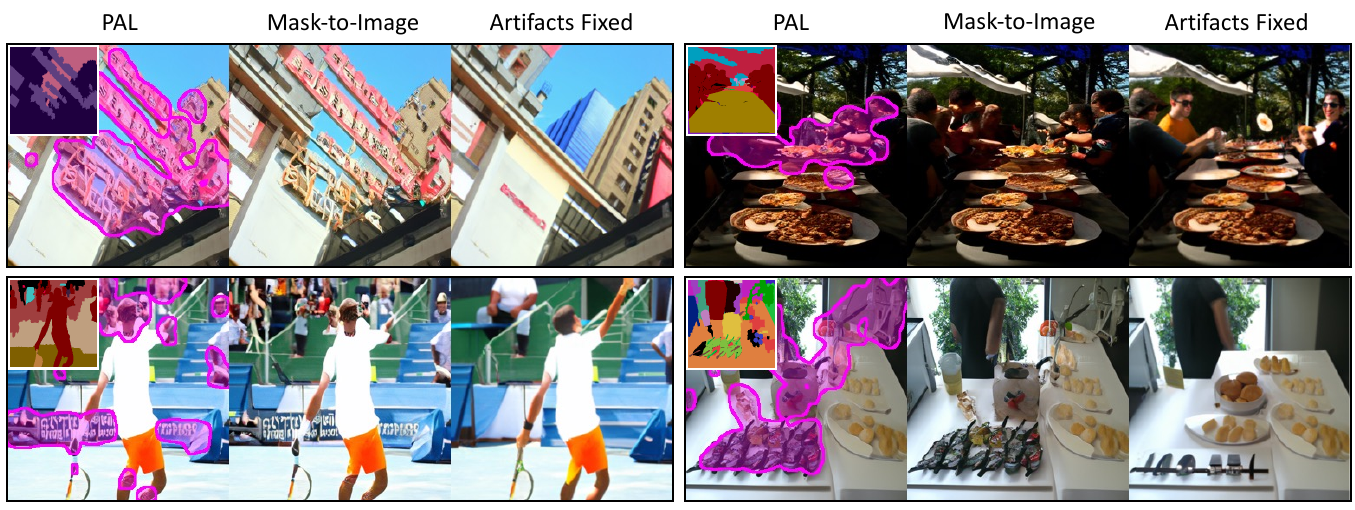}
    \vspace{-20 pt}
    \caption{More qualitative results for perceptual artifacts localization (PAL) prediction and the artifacts fixed images for Mask-to-Image translation with PITI \cite{wang2022pretraining}.}
    \label{fig:supp_mask2image_fix}
    \vspace{-5 pt}
\end{figure*}

\begin{figure*}[!h]
 \vspace{-10 pt}
    \centering
    \includegraphics[trim=0in 5.7in 4.3in 0in, clip,width=\textwidth]{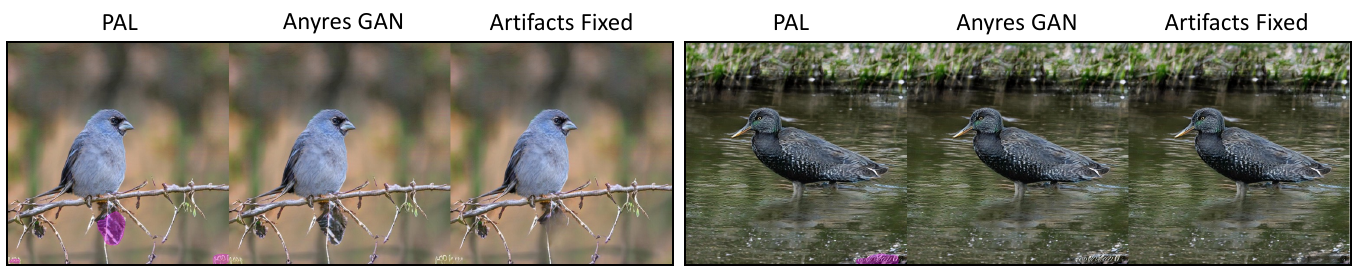}
    \vspace{-20 pt}
    \caption{More qualitative results for perceptual artifacts localization (PAL) prediction and the artifacts fixed images for Anyres GAN \cite{chai2022any}. }
    \label{fig:supp_anyresgan_fix}
    \vspace{-5 pt}
\end{figure*}

\begin{figure*}[!h]
 \vspace{-10 pt}
    \centering
    \includegraphics[trim=0in 3.4in 4.3in 0in, clip,width=\textwidth]{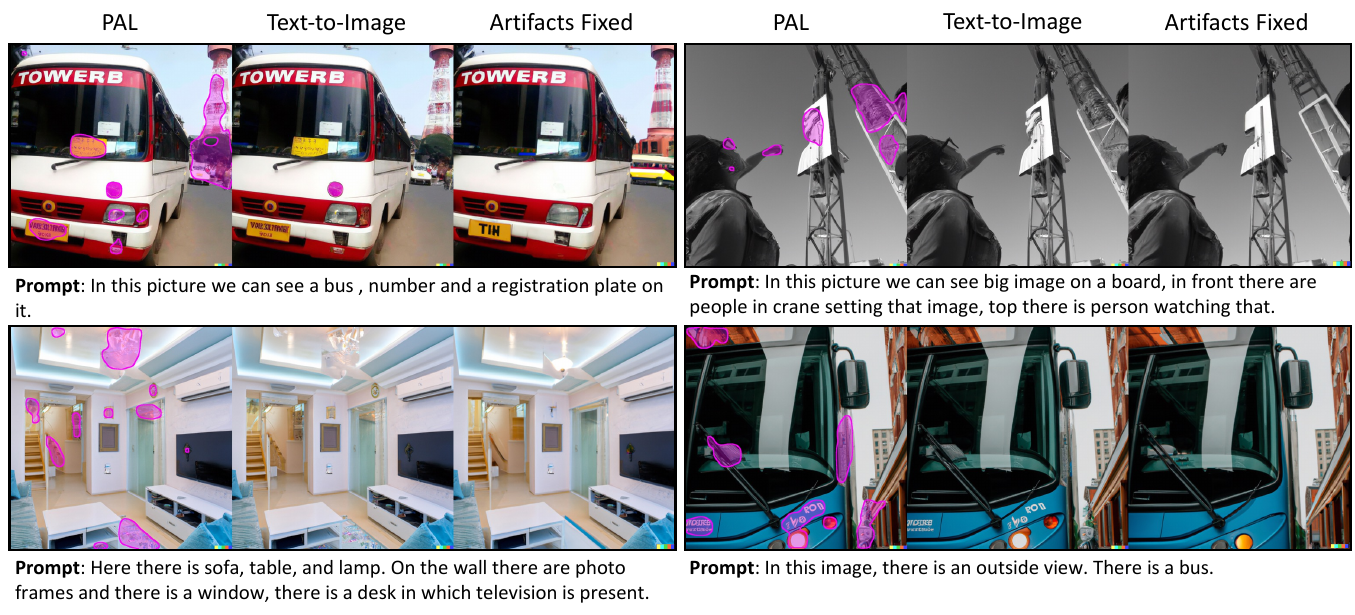}
    \vspace{-20 pt}
    \caption{More qualitative results for perceptual artifacts localization (PAL) prediction and the artifacts fixed images for Text-to-Image synthesis with DALL-E 2 \cite{ramesh2022hierarchical}.}
    \label{fig:supp_text2image_fix}
    \vspace{-5 pt}
\end{figure*}

\begin{figure*}[!h]
    \centering
    \includegraphics[trim=0in 3.5in 3.7in 0in, clip,width=\textwidth]{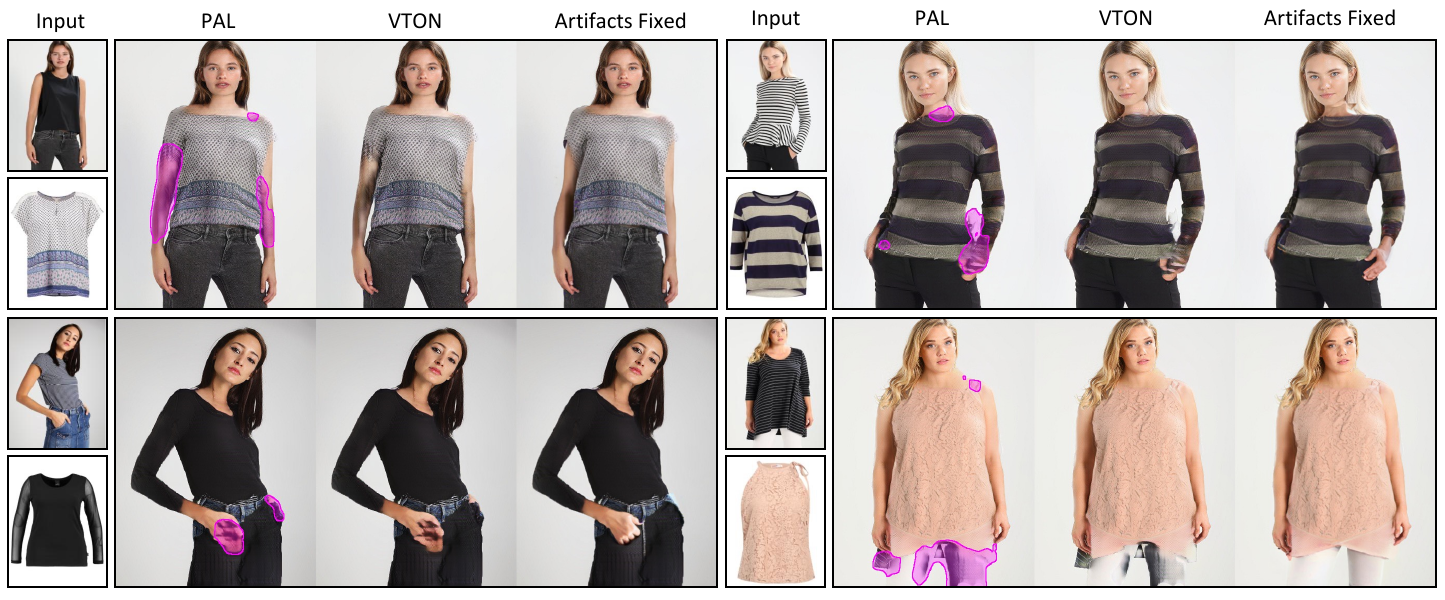}
    \vspace{-20 pt}
    \caption{More qualitative results for perceptual artifacts localization (PAL) prediction and the artifacts fixed images for virtual try-on with \cite{fele2022c}.}
    \label{fig:supp_vton_fix}
\end{figure*}

\begin{figure*}[!h]
 \vspace{-10 pt}
    \centering
    \includegraphics[trim=0in 4.1in 4.3in 0in, clip,width=\textwidth]{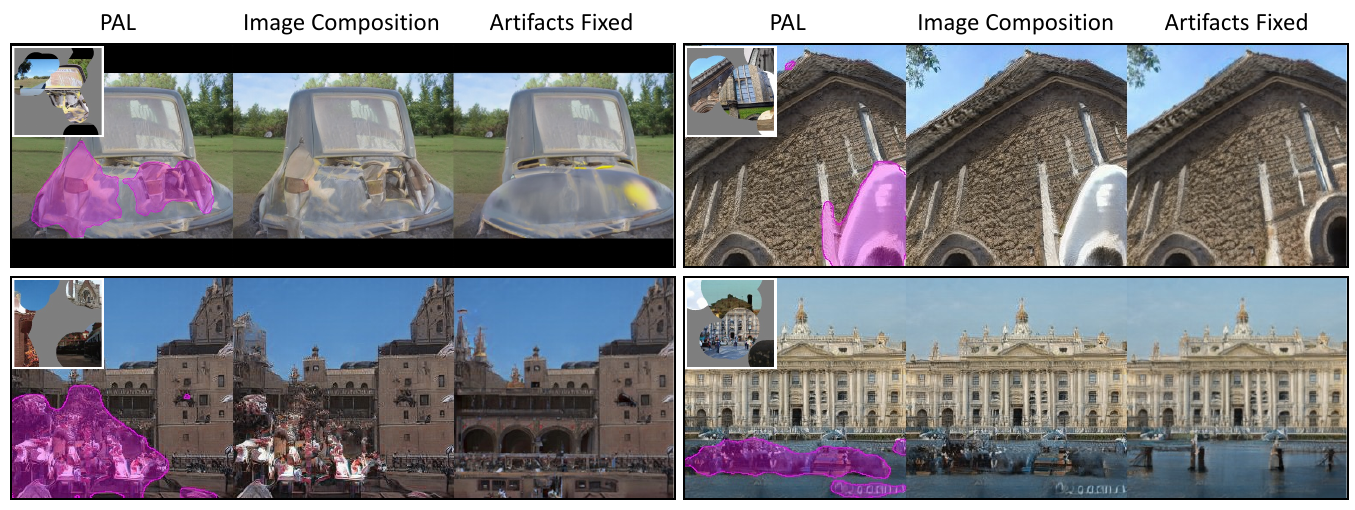}
    \vspace{-20 pt}
    \caption{More qualitative results for perceptual artifacts localization (PAL) prediction and the artifacts fixed images for latent composition \cite{chai2021using}.}
    \label{fig:supp_composition_fix}
    \vspace{-5 pt}
\end{figure*}

\begin{figure*}[!h]
 \vspace{-10 pt}
    \centering
    \includegraphics[trim=0in 5.7in 4.3in 0in, clip,width=\textwidth]{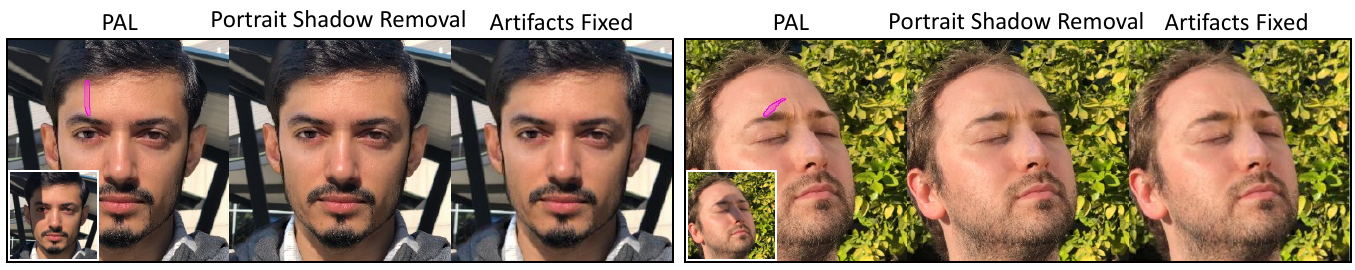}
    \vspace{-20 pt}
    \caption{More qualitative results for perceptual artifacts localization (PAL) prediction and the artifacts fixed images for Portrait Shadow Removal \cite{zhang2020portrait}. Please \emph{zoom in} to see the detailed comparisons. }
    \label{fig:supp_psr_fix}
    \vspace{-5 pt}
\end{figure*}

We show more qualitative results of perceptual artifacts segmentation and artifacts fixed results for ten synthesis tasks. These visual results are shown in Fig \ref{fig:supp_stylegan_fix} - \ref{fig:supp_psr_fix}. In each example, first image is the generated image with perceptual artifacts localization (PAL), which is indicated by the pink mask. The second image is the original generated image, and the third is the corresponding artifacts fixed image using the predicted PAL. We put the original and artifacts refined images side-by-side for more direct visual comparison. 


\subsection{The Choices of Inpainting Models}

\begin{figure*}[!t]
 \vspace{-10 pt}
    \centering
    \includegraphics[trim=0in 1.0in 6.6in 0in, clip,width=\textwidth]{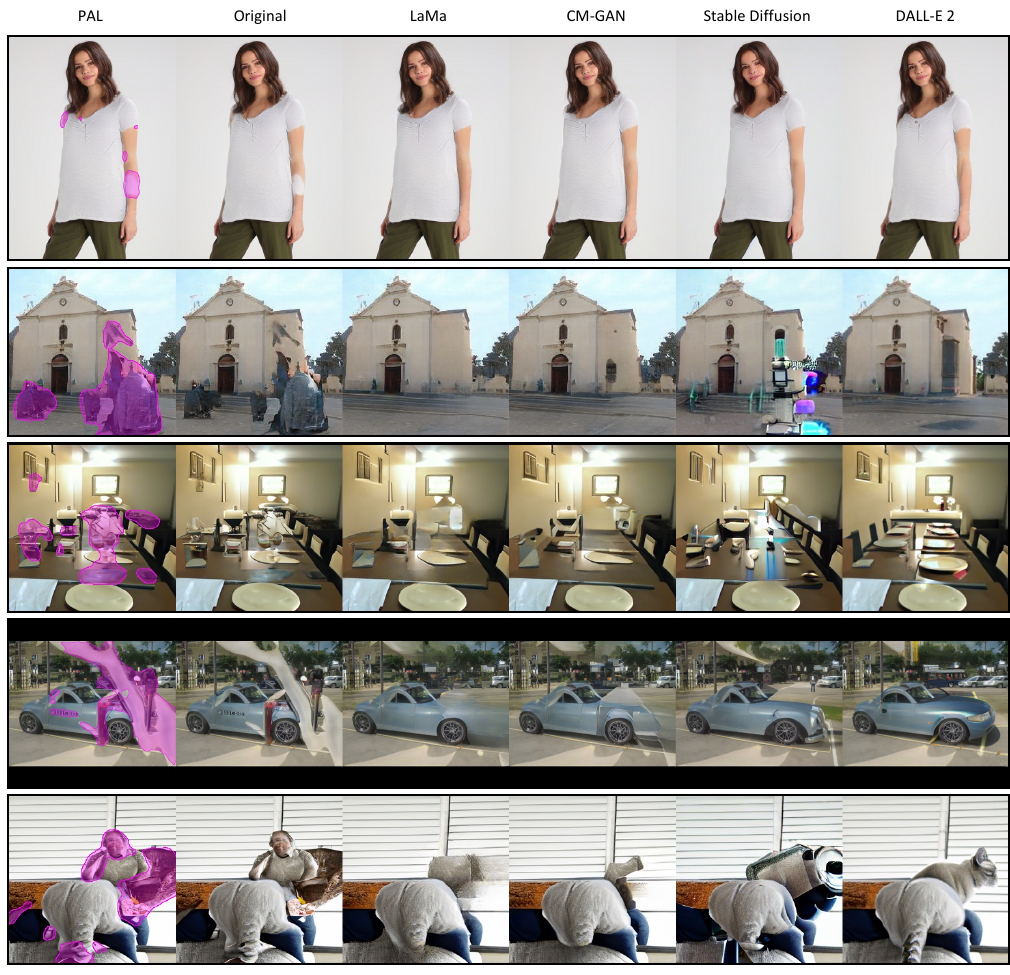}
    \vspace{-20 pt}
    \caption{An ablation study on how four different state-of-the-arts inpainting models, including LaMa \cite{suvorov2022resolution}, CM-GAM \cite{zhang2022inpainting}, Stable Duffion \cite{rombach2022high}, and DALL-E 2\cite{ramesh2022hierarchical}, could fix the perceptual artifacts in types of generated images using our PAL prediction as the inpainting masks. }
    \label{fig:supp_inpaint_comparison}
    \vspace{-5 pt}
\end{figure*}

\begin{figure*}[!h]
 \vspace{-10 pt}
    \centering
    \includegraphics[trim=0in 5.3in 1.6in 0in, clip,width=\textwidth]{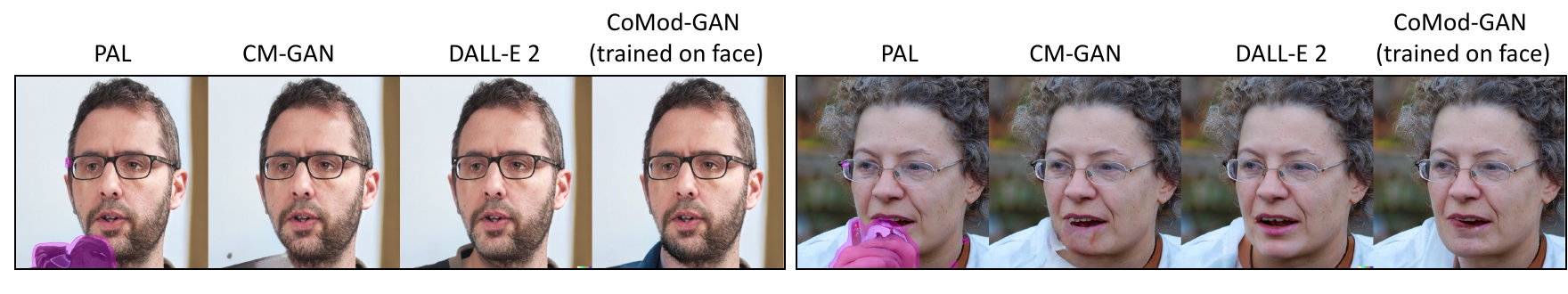}
    \vspace{-20 pt}
    \caption{An ablation study on how inpainting models work on face artifacts removal. Note that CM-GAN \cite{zheng2022image} and DALL-E 2 \cite{ramesh2022hierarchical} are not tailored for face inpainting, while CoMod-GAN \cite{zhao2021large} is trained on the FFHQ \cite{karras2019style} dataset for face inpainting specifically. }
    \label{fig:supp_face_inpaint_comparison}
    \vspace{-5 pt}
\end{figure*}

In this paper, we mainly use CoMod-GAN \cite{zhao2021large}, LaMa \cite{suvorov2022resolution}, and DALL-E 2 inpainting \cite{ramesh2022hierarchical} in our artifacts fixing pipeline, as discussed in section 4 in the paper. In this section, we show ablation studies on how different inpainting models can be used to fix the perceptual artifacts in different cases. As shown in Figure \ref{fig:supp_face_inpaint_comparison}, for face inpainting, CoMod-GAN trained on FFHQ \cite{karras2019style} face dataset produce more realistic results than the CM-GAN \cite{zheng2022image}, and has similar performance to DALL-E 2 inpainting. Since CoMod-GAN has faster inference speed than DALL-E 2 by a order of magnitude, we choose CoMod-GAN for general face inpainting cases. For other in-the-wild inpainting cases, as shown in Figure \ref{fig:supp_inpaint_comparison}, we observe that GAN-based models LaMa and CM-GAN have reasonably good performance on the relatively easy cases, such as the first two rows. However, when the images are under perspective ($3^{rd}$ row) or involve object completion ($4^{th}$ and $5^{th}$ rows), diffusion-based models generally produce much better results. Within diffusion-based models, DALL-E 2 produce much more realistic details than Stable Diffusion inpainting \cite{rombach2022high} with v1.5 checkpoint. Therefore, we use LaMa for the easy background inpainting in tasks like Anyres GAN \cite{chai2022any}, and DALL-E 2 for the rest of tasks with complex scene or object completion. 


\subsection{Zoom-in Effect on Inpainting}

\begin{figure*}[!t]
 \vspace{-10 pt}
    \centering
    \includegraphics[trim=0in 0.0in 2.1in 0in, clip,width=\textwidth]{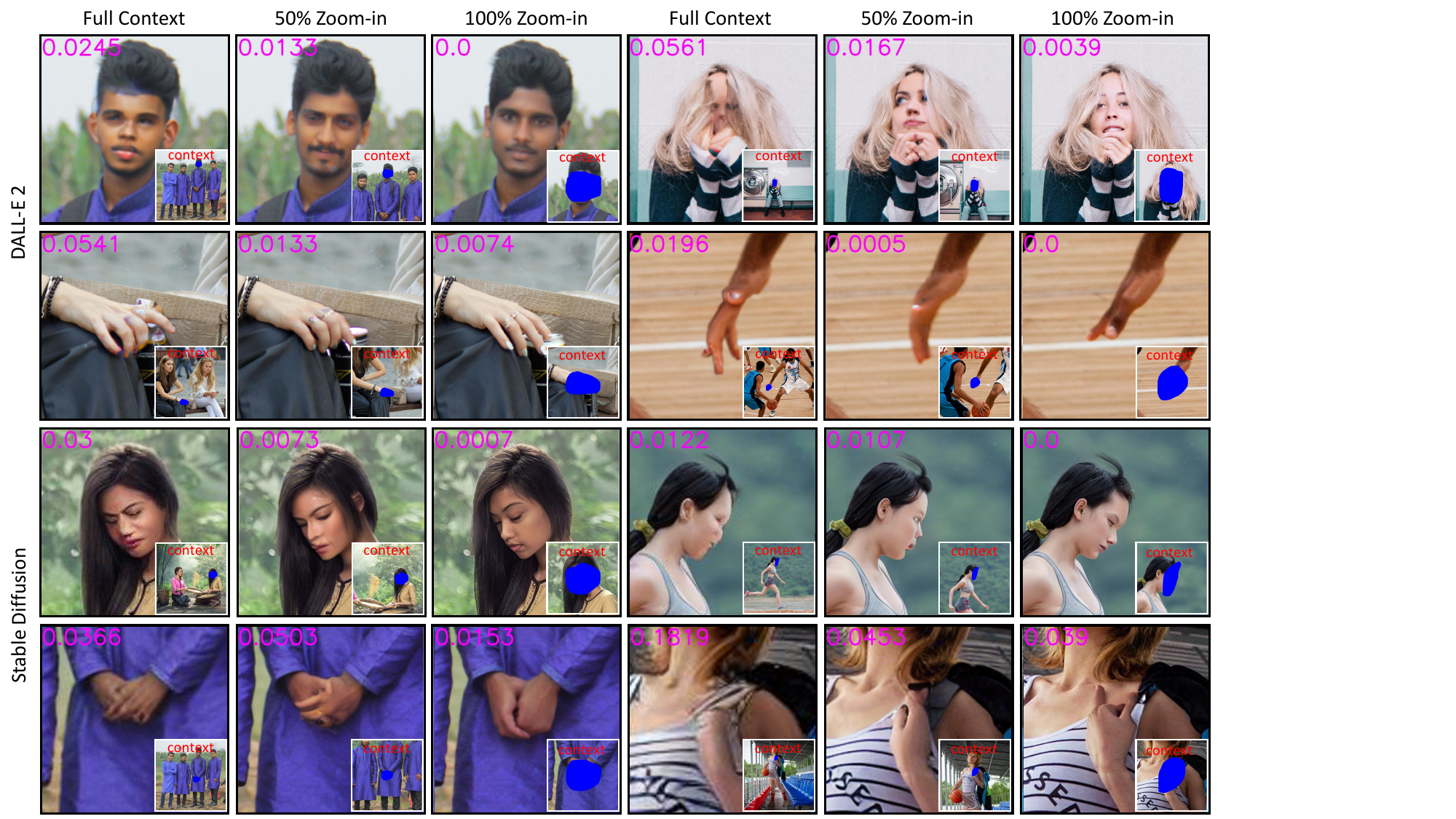}
    \vspace{-10 pt}
    \caption{More qualitative results showing that DALL-E 2 inpainting \cite{ramesh2022hierarchical} and Stable Diffusion \cite{ramesh2022hierarchical} tend to generate less perceptual artifacts when zooming in around the object region, such as faces and hands. We show that our PAR scores, which are placed at the top left corner of the images, can be used to quantify this observation and confirm our insight. }
    \label{fig:supp_dalle_bias}
    \vspace{-5 pt}
\end{figure*}

\begin{figure*}[!t]
 \vspace{-10 pt}
    \centering
    \includegraphics[trim=0in 4.7in 0.65in 0in, clip,width=\textwidth]{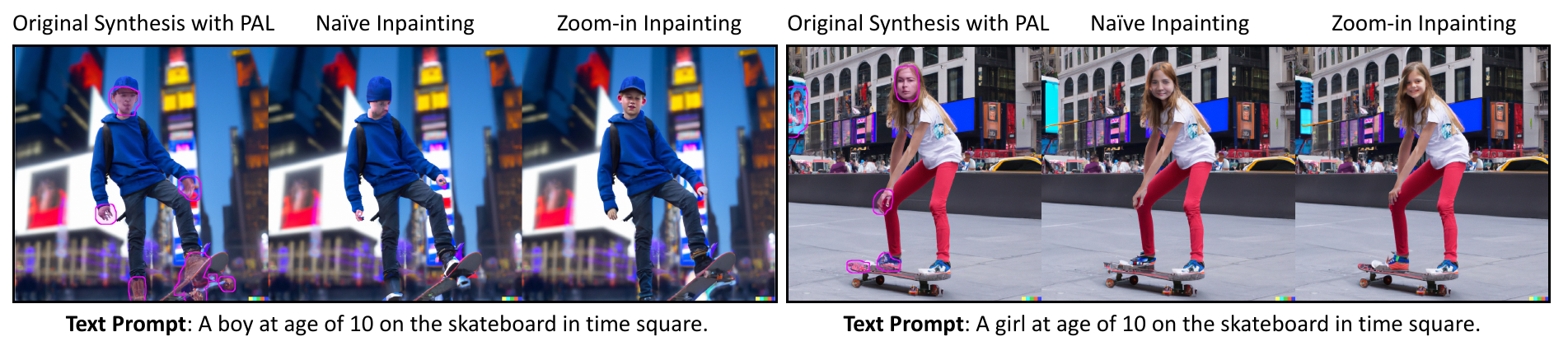}
    \vspace{-10 pt}
    \caption{Qualitative comparison between naive inpainting and zoom-in inpainting for fixing perceptual artifacts in text-to-image outputs. In the above examples, we use DALL-E 2 \cite{ramesh2022hierarchical} for both text-to-image generation and inpainting. Naive inpainting could fix certain artifacts compared to the original synthesis, but still struggles to generate high-fidelity object details. In contrast, zoom-in inpainting pipeline produces much more realistic object details. }
    \label{fig:supp_zoom_in_fix}
    \vspace{-5 pt}
\end{figure*}

In the main paper, we discuss that diffusion-based models, i.e. DALL-E 2 \cite{ramesh2022hierarchical}, systematically struggles to generate high-fidelity object details, such as faces and hands. Here, we show more qualitative results. Inspired by this insight, we further propose a 'zoom-in' inpainting pipeline that can fix the perceptual artifacts in the object detail level. As show in Figure \ref{fig:supp_zoom_in_fix}, we can see that this zoom-in inpainting pipeline can significantly refine the object details and outperform naively inpainting using the full images and masks. More detailed comparisons on hands and faces are illustrated in Figure \ref{fig:supp_dalle_bias}. In this work, we use the fixed text prompt for all the patches, but more tailored text prompt for the individual cropped patch should theoretically improve the visual quality, which we leave as future work.

\section{SDEdit for Perceptual Artifacts Fixing}

\begin{figure}[!h]
    \centering
    \includegraphics[trim=0in 1.5in 5.8in 0in, clip,width=0.47\textwidth]{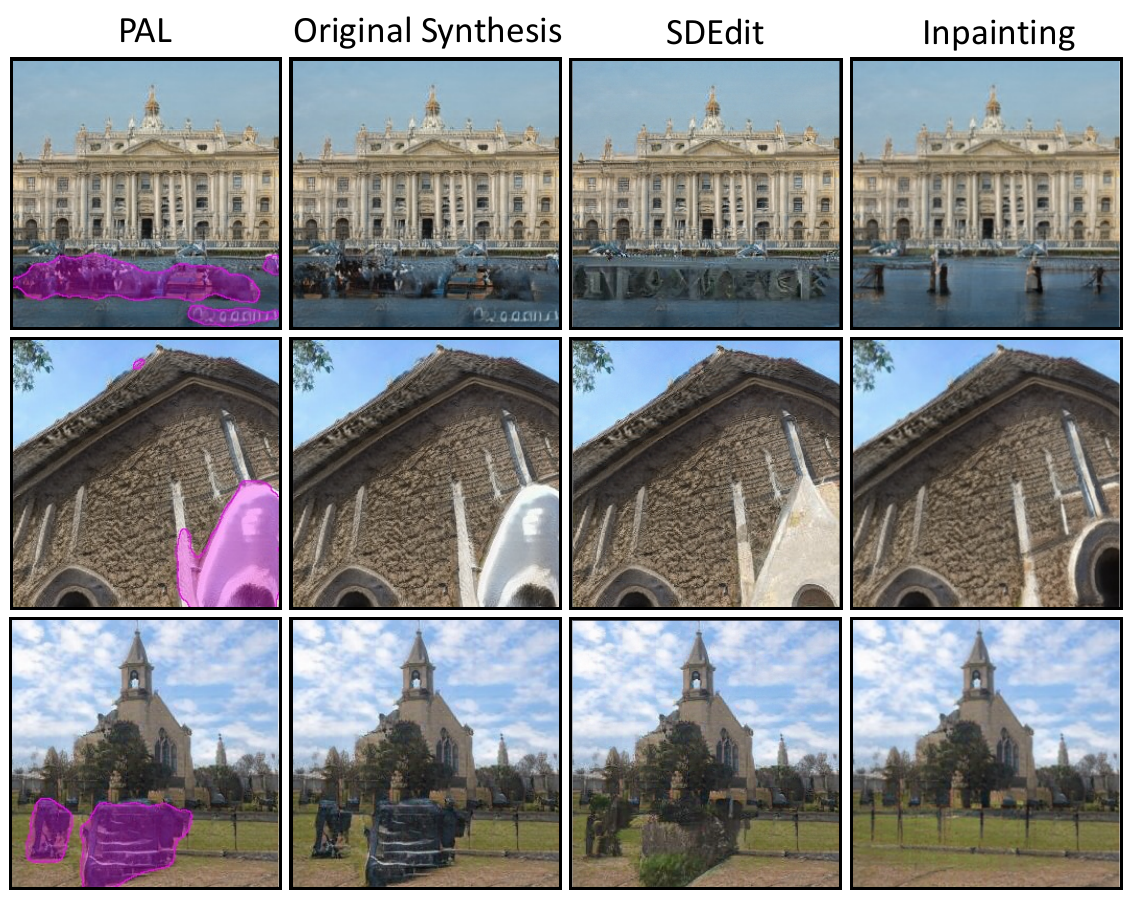}
    \caption{Qualitative comparison between SDEdit \cite{meng2021sdedit} and DALL-E 2 inpainting \cite{ramesh2022hierarchical} for artifacts fixing. In general, we can see that SDEdit preserves more image identity (more similar to the original synthesis), while DALL-E 2 inpainting produces better realism. }
    \label{fig:supp_sdedit}
\end{figure}

Using inpainting methods to fix the perceptual artifacts might not be ideal for certain synthesis tasks, since it could change too much of the original generated image identity. We also explore an alternative approach SDEdit \cite{meng2021sdedit}, which enables stroke-based editing using a diffusion model generative prior DDIM \cite{song2020denoising}. In the implementation, we convert the pixels in the perceptual artifacts region into stroke painting by RTV smooth algorithm \cite{xu2012structure}, and then run SDEdit to re-generate pixels in the artifacts region. As shown in Figure \ref{fig:supp_sdedit}, SDEdit preserves more image identity with respect to the original generation, but underperforms DALL-E 2 inpainting \cite{ramesh2022hierarchical} in terms of realism. SDEdit also has a hyperparameter that controls the tradeoff between realism and faithfulness (identity preservation), and this can be adjusted for different tasks. In this work, we showcase the usage of SDEdit with DDIM trained on LSUN Church dataset. To apply this in the wild, we might either need to re-train DDIM in larger diverse dataset or integrate SDEdit with other diffusion-based models, i.e. Stable Diffusion \cite{rombach2022high}, and we leave this as future work. 

\end{document}